\documentclass[journal]{IEEEtran}

\usepackage{graphicx}
\usepackage{amssymb,amsmath,epsfig}
\usepackage{algorithm}
\usepackage{algorithmic}
\usepackage{booktabs}
\usepackage{mathrsfs}
\usepackage{color}
\usepackage{textcomp}
\usepackage{bbding}
\usepackage{pifont}
\usepackage{wasysym}
\usepackage{amssymb}
\usepackage{subcaption}
\usepackage{cite}
\usepackage{amsmath,amssymb,amsfonts}
\usepackage{xcolor}
\usepackage{multirow}
\usepackage{epsfig}
\usepackage{adjustbox}
\usepackage[colorlinks,
            linkcolor=red,
            anchorcolor=blue,
            citecolor=green
            ]{hyperref}

\begin{document}
%
\title{Cross-receptive Focused Inference Network for Lightweight Image Super-Resolution}
%
%
%
\author{Wenjie Li,
        Juncheng Li,
        Guangwei Gao,~\IEEEmembership{Senior Member,~IEEE,}
        Weihong Deng,~\IEEEmembership{Member,~IEEE,}
        Jiantao Zhou,~\IEEEmembership{Senior Member,~IEEE,}
        Jian Yang,~\IEEEmembership{Member,~IEEE}
        and Guo-Jun Qi,~\IEEEmembership{Fellow,~IEEE}
\thanks{This work was supported in part by the National Natural Science Foundation of China under Grant 61972212 and 61833011, the Six Talent Peaks Project in Jiangsu Province under Grant RJFW-011, the Shanghai Sailing Program under Grant 23YF1412800, the Natural Science Foundation of Shanghai under Grant 23ZR1422200, and the Open Fund Project of Provincial Key Laboratory for Computer Information Processing Technology (Soochow University) under Grant KJS2274.~\textit{(Corresponding author: Guangwei Gao.)}}
\thanks{Wenjie Li and Guangwei Gao are with the Intelligent Visual Information Perception Laboratory, Institute of Advanced Technology, Nanjing University of Posts and Telecommunications, Nanjing 210046, China, and also with the Provincial Key Laboratory for Computer Information Processing Technology, Soochow University, Suzhou 215006, China (e-mail: csggao@gmail.com,lewj2408@gmail.com).}
\thanks{Juncheng Li is with the School of Communication and Information Engineering, Shanghai University, Shanghai 200444, China, and also with Jiangsu Key Laboratory of Image and Video Understanding for Social Safety, Nanjing University of Science and Technology, Nanjing 210094, China (e-mail: cvjunchengli@gmail.com).}
\thanks{Weihong Deng is with the Pattern Recognition and Intelligent System Laboratory, School of Artificial Intelligence, Beijing University of Posts and Telecommunications, Beijing 100876, China (e-mail: whdeng@bupt.edu.cn).}
\thanks{Jiantao Zhou is with the State Key Laboratory of Internet of Things for Smart City, Department of Computer and Information Science, Faculty of Science and Technology, University of Macau, Macau 999078, China (e-mail: jtzhou@um.edu.mo).}
\thanks{Jian Yang is with the School of Computer Science and Technology, Nanjing University of Science and Technology, Nanjing 210094, China (e-mail: csjyang@njust.edu.cn).}
\thanks{Guo-Jun Qi is with the Research Center for Industries of the Future and the School of Engineering, Westlake University, Hangzhou 310024, China, and also with OPPO Research, Seattle, WA 98101 USA (e-mail: guojunq@gmail.com).}

}

\markboth{IEEE Transactions on Multimedia}%
{Shell \MakeLowercase{\textit{et al.}}: Bare Demo of IEEEtran.cls for IEEE Journals}
%

\maketitle

\begin{abstract}
Recently, Transformer-based methods have shown impressive performance in single image super-resolution (SISR) tasks due to the ability of global feature extraction. However, the capabilities of Transformers that need to incorporate contextual information to extract features dynamically are neglected. To address this issue, we propose a lightweight Cross-receptive Focused Inference Network (CFIN) that consists of a cascade of CT Blocks mixed with CNN and Transformer. Specifically, in the CT block, we first propose a CNN-based Cross-Scale Information Aggregation Module (CIAM) to enable the model to better focus on potentially helpful information to improve the efficiency of the Transformer phase. Then, we design a novel Cross-receptive Field Guided Transformer (CFGT) to enable the selection of contextual information required for reconstruction by using a modulated convolutional kernel that understands the current semantic information and exploits the information interaction within different self-attention. Extensive experiments have shown that our proposed CFIN can effectively reconstruct images using contextual information, and it can strike a good balance between computational cost and model performance as an efficient model. Source codes will be available at \textit{\url{https://github.com/IVIPLab/CFIN}}.
\end{abstract}


%
\IEEEpeerreviewmaketitle

\section{Introduction}
\begin{figure}[t]
\centering
\includegraphics[width=0.45\textwidth,trim=0 0 40 40]{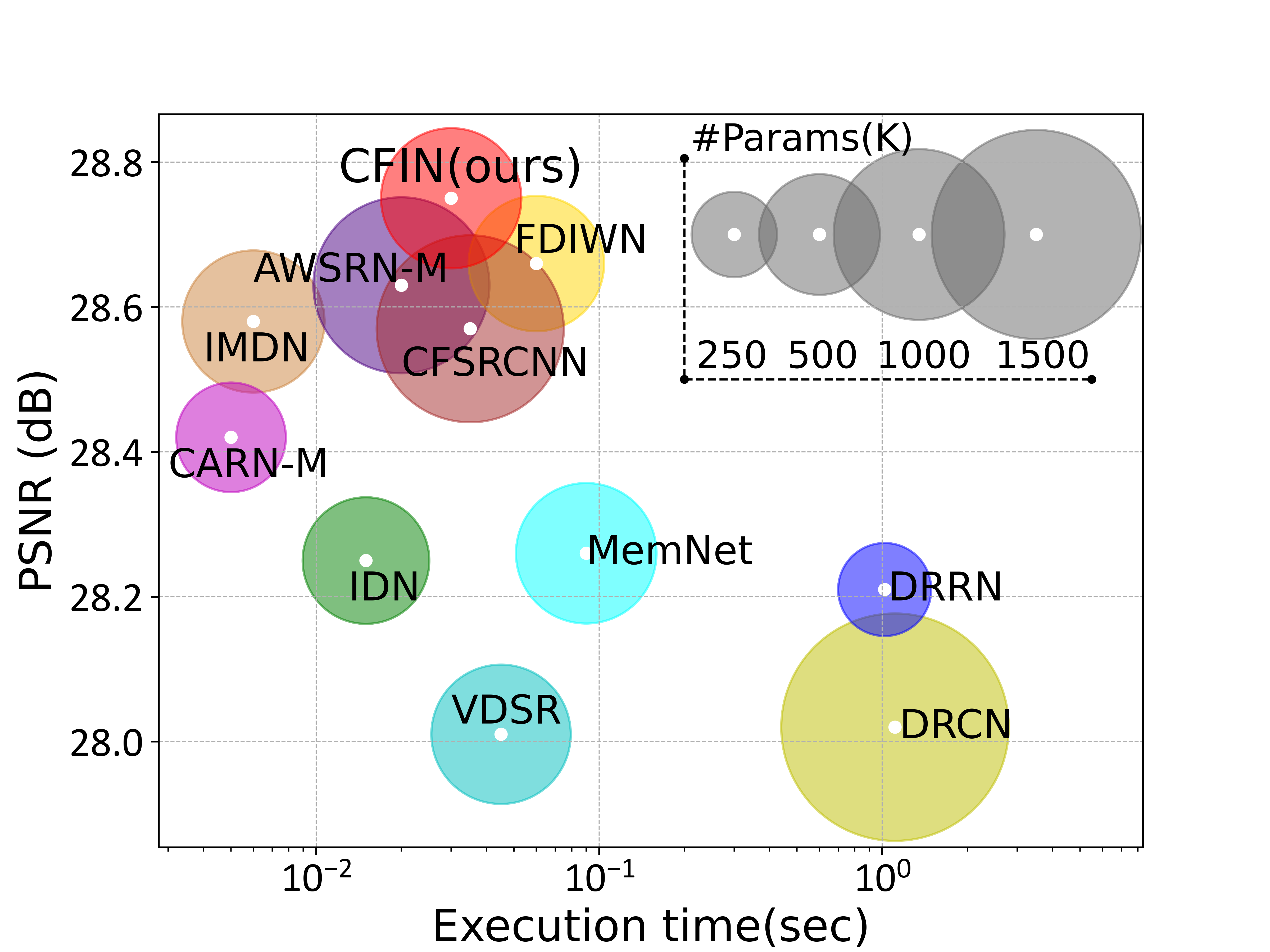}
\caption{Model inference time studies on Set14 ($ \times 4$).}
\label{Time}
\end{figure}

The task of Single Image Super-Resolution (SISR) aims to estimate a realistic High-Resolution (HR) image from the Low-Resolution (LR) one, which plays a fundamental role in various computer vision tasks, including face imaging~\cite{jiang2020dual,gao2023ctcnet}, hyperspectral imagery~\cite{li2022symmetrical}, video processing~\cite{hu2022you}, and medical imaging~\cite{li2022transformer}. As an ill-posed problem, SISR is still a challenging task. To solve this task, many Convolutional Neural Networks (CNN) based methods have been proposed to directly learn the mapping between the LR and HR image pairs. For example, Dong \textit{et al.} presented the first CNN-based model, dubbed SRCNN~\cite{dong2015image}. Although SRCNN only has three convolutional layers, its performance is significantly better than traditional solutions. Subsequently, a series of networks with complex architectures were proposed, and deep CNN-based methods have achieved remarkable progress in SISR~\cite{zhang2018residual}. Although these models have achieved promising results, their computational cost is often too huge (Fig.~\ref{Time}) to be popularized and widely used.

To solve the problems mentioned above, constructing lightweight SISR models has attracted more and more attention. Among these CNN-based lightweight models, most of them focus on efficient network architecture design, such as neural architecture search~\cite{chu2020multi}, multi-scale structure design~\cite{li2018multi,li2020mdcn}, and channel grouping strategy~\cite{ahn2018fast}. However, convolution kernels can only extract local features, which is difficult to model the long-term dependencies of the image. As a complementary solution to CNN, Transformer has achieved excellent performance in many visual tasks with its powerful global modeling capability~\cite{zamir2021restormer,Yu2022CVPR}. Recently, some Transformer-based SISR methods have been proposed~\cite{gao2022lightweight,chen2022activating}. For instance, SwinIR~\cite{liang2021swinir} introduced Transformer into SISR, relying on its advantage of using a shifted window scheme to model long-term dependencies, showing the great promise of Transformer in SISR. ESRT~\cite{lu2021efficient} is an efficient SISR model that combines a lightweight CNN and lightweight Transformer in an end-to-end model. However, most existing Transformer-based methods ignore the importance of dynamic modeling with context. As studied in neurology~\cite{gilbert2013top}, the morphology of neurons should change adaptively with changes in the environment. This adjustment mechanism has been studied in many fields. For example, Jia \textit{et al.}~\cite{jia2016dynamic} generated convolution kernel weights from the features extracted from another network. Chen \textit{et al.}~\cite{chen2020dynamic} aggregated multiple convolution kernels in parallel and based on local attention to adaptively adjust the weights. Lin \textit{et al.}~\cite{lin2020context} presented the context-gated convolution to incorporate contextual awareness into the convolutional layer. It is worth noting that each pixel in an image cannot be isolated, they should have some relationship to those pixels around it, and this relationship is referred as contextual information in the text. Such a designation has been widely used in image segmentation tasks~\cite{yuan2020object}.

Motivated by the above methods, in this paper, we introduce the power of contextual reasoning into Transformer and propose a lightweight Cross-receptive Focused Inference Network (CFIN) for SISR. CFIN is a hybrid network composed of a Convolutional Neural Network (CNN) and a Transformer. In the convolution stage, a Cross-scale Information Aggregation Module (CIAM) is designed to extract more potentially useful information with the help of the Redundant Information Filter Unit (RIFU). In the Transformer stage, we propose a Cross-receptive Field Guide Transformer (CFGT) to achieve cross-scale long-distance information fusion with specially designed Context Guided Attention (CGA). In summary, the main contributions of this paper can be summarized as follows:
\begin{itemize}
  \item We propose a Redundant Information Filter Unit (RIFU), which can remove redundant information and learn flexible local features. Meanwhile, an efficient Cross-scale Information Aggregation Module (CIAM) is specially designed for elaborately combining several RIFUs to ensure full use of local features.
  \item We propose a Context Guided Attention (CGA) scheme, which can adaptively adjust the weights of the modulating convolution kernel to achieve the selection of the desired contextual information. In addition, a novel Cross-receptive Field Guide Transformer (CFGT) is proposed to combine CGAs of different receptive fields to further facilitate contextual interactions.
  \item We propose a lightweight Cross-receptive Focused Inference Network (CFIN) for SISR. CFIN elegantly integrates CNN and Transformer, achieving a good balance between computational cost and model performance.
\end{itemize}


\section{Related Work}~\label{RW}
\subsection{Lightweight SISR Model}
Due to the powerful learning ability of neural networks, more and more effective SISR methods based on neural networks are proposed~\cite{jiang2020hierarchical,he2021towards,tang2021bridgenet,wu2022bridging,li2022learning}. However, most of the methods are limited to real-world applications due to their huge computational cost. To handle this issue, some lightweight and efficient SISR methods based on model architecture have been presented. For example, IDN~\cite{hui2018fast} used an information distillation network to selectively fuse features, and then IMDN~\cite{hui2019lightweight} improved it to build a lighter and faster model. RFDN~\cite{liu2020residual} combined channel splitting and residual structure to achieve better performance. LatticeNet~\cite{luo2020latticenet} enhanced the representation of the model with designed lattice blocks. PFFN~\cite{zhang2021pffn} made full use of the feature map for each layer by the proposed pixel attention. FDIWN~\cite{gao2022feature} improved the model performance by fully using the intermediate layer features. LatticeNet-CL~\cite{luo2022lattice} further enhanced LatticeNet by using contrast loss as a regularization constraint. In addition, some recent Transformer-based approaches have made promising progress in SISR tasks. SwinIR~\cite{liang2021swinir} performed global attention operations separately on the divided windows and achieved performance beyond the CNN model. ESRT~\cite{lu2021efficient} and LBNet~\cite{gao2022lightweight} performed feature splitting by efficient multi-headed attention, significantly reducing the training memory of the Transformer. However, existing lightweight methods often neglect the importance of contextual information, which may be useful for image reconstruction.



\subsection{Context Reasoning}
With an in-depth understanding of deep learning, researchers tentatively explored how to increase the contextual information of the model. It can be roughly divided into the following categories. For example, using the attention mechanism to modify the feature representation, and the typical one is to modify the local features through the attention mechanism~\cite{hu2018squeeze}. However, most of them can only modify the features by changing the input mapping. Recently, some works~\cite{wu2019pay,zhu2019deformable,chen2020dynamic,jo2018deep} have tried to dynamically change network parameters by analyzing local or global information. However, they only consider local fragments~\cite{wu2019pay}, ignore weight tensors in convolutional layers~\cite{zhu2019deformable}, or have high training costs~\cite{jo2018deep}. Researchers in~\cite{yang2018convolutional} imitated the human visual system and simulated the bottom-up impact of semantic information on the model through reverse connections, but this feedback mechanism is difficult to explain in the model. In addition, none of them use contextual information to guide global attention interactions dynamically. In this work, we aim to introduce context reasoning to further enhance the performance of the model and build a model that can adaptively modify the network weights.

\begin{figure*}
\begin{center}
\includegraphics[width=0.8\linewidth]{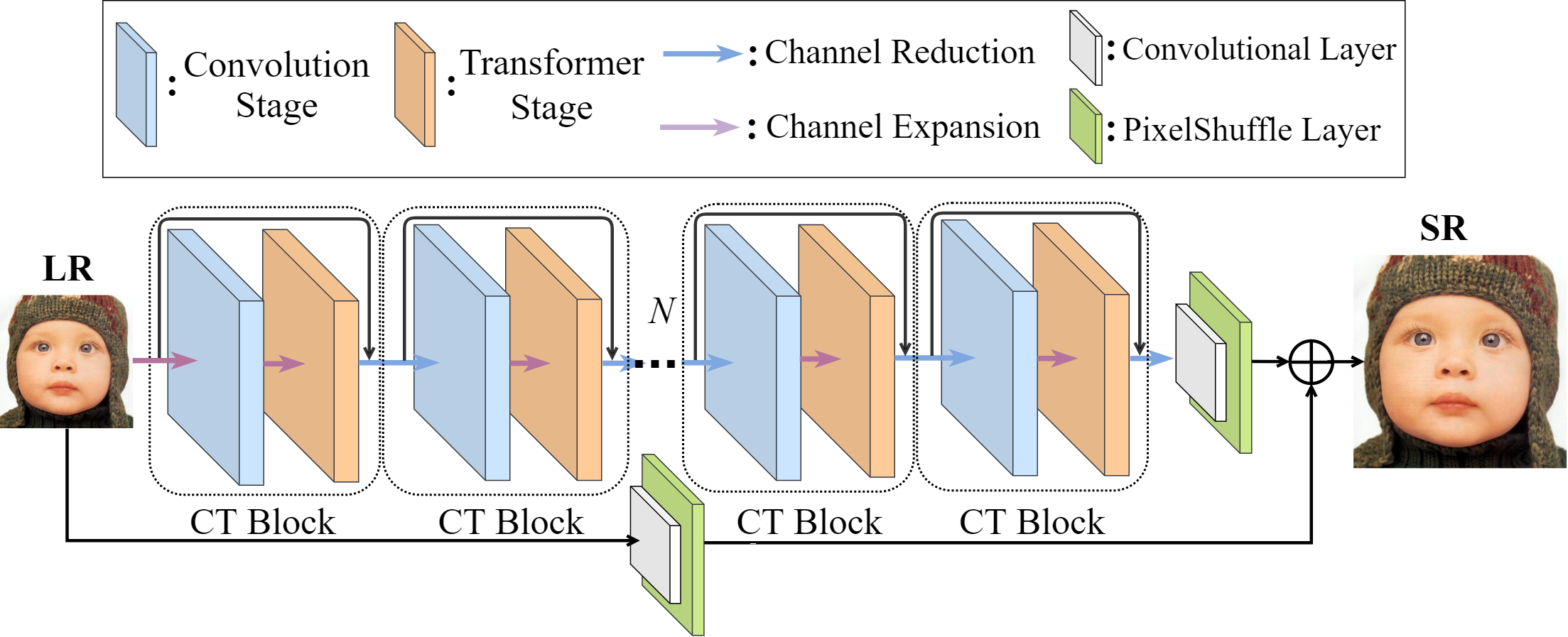}
\end{center}
   \caption{The overall architecture of the proposed Cross-receptive Focused Inference Network (CFIN). It is worth noting that $1 \times 1$ convolution is used for channel reduction and channel expansion in CFIN.}
\label{main}
\end{figure*}

\section{Proposed Method}~\label{ME}
\subsection{Cross-receptive Focused Inference Network}
In this paper, we devise a lightweight Cross-receptive Focused Inference Network (CFIN) for SISR. As shown in Fig.~\ref{main}, CFIN consists of several CNN-Transformer (CT) blocks and two PixelShuffle layers. Meanwhile, each CT block contains a convolution stage and a Transformer stage. And each two CT block is called once using a loop mechanism for a better trade-off between model size and performance. We define the input and output of CFIN as ${I_{LR}}$ and ${I_{SR}}$, respectively. Firstly, the dimension of the input image is rapidly increased to obtain shallow features $I_{shallow}$ for subsequent processing
\begin{equation}
I_{shallow} = F_{ce}(I_{LR}),
\end{equation}
where $F_{ce}(\cdot)$ is the channel expansion operation. Then, the shallow features are sent to CT blocks for feature extraction, and the complete operation of each CT block can be defined as follows
\begin{equation}
I_{ct}^{i} = F^{i}_{CT}(I_{in}^{i}) = F_{cr}(F_{T}(F_{ce}(F_{C}(I_{in}^{i})))) + I_{in}^{i},
\end{equation}
    where $F_{cr}(\cdot)$, $F_{ce}(\cdot)$, $F_{C}(\cdot)$, and $F_{T}(\cdot)$ denote the channel reduction operation, channel expansion, the convolution stage, and the Transformer stage, respectively. $I_{in}^{i}$ and $I_{ct}^{i}$ are the input and output of the $i$-th CT block. After the operation of $N$ CT blocks, we can attain the final deep features as
\begin{equation}
I_{ct} = F^{N}_{CT}(F^{N-1}_{CT}(\cdots F^{2}_{CT}(F^{1}_{CT}(I_{shallow})))),
\end{equation}
where $I_{ct}$ denotes the output of the $N$-th CT block. Finally, to obtain the final SR image, both $I_{ct}$ and $I_{LR}$ are simultaneously fed into the post-sampling reconstruction module
\begin{equation}
I_{SR} = F_{rec}(I_{ct}) + F_{rec}(I_{LR}),
\end{equation}
where $F_{rec}(\cdot)$ is the post-sampling reconstruction module, which is composed of a $3 \times 3$ convolutional layer and a PixelShuffle layer.

\subsection{Convolution Stage}
In the convolution stage, we propose a Cross-scale Information Aggregation Module (CIAM) to refine potential image information and make the model understand the preliminary SR information. As shown in Fig.~\ref{CIAM}, CIAM is mainly composed of three Redundant Information Filter Units (RIFU).

\subsubsection{Redundant Information Filter Unit (RIFU)}
\begin{figure}[t]
\centering
\includegraphics[width=0.45\textwidth,trim=0 0 20 0]{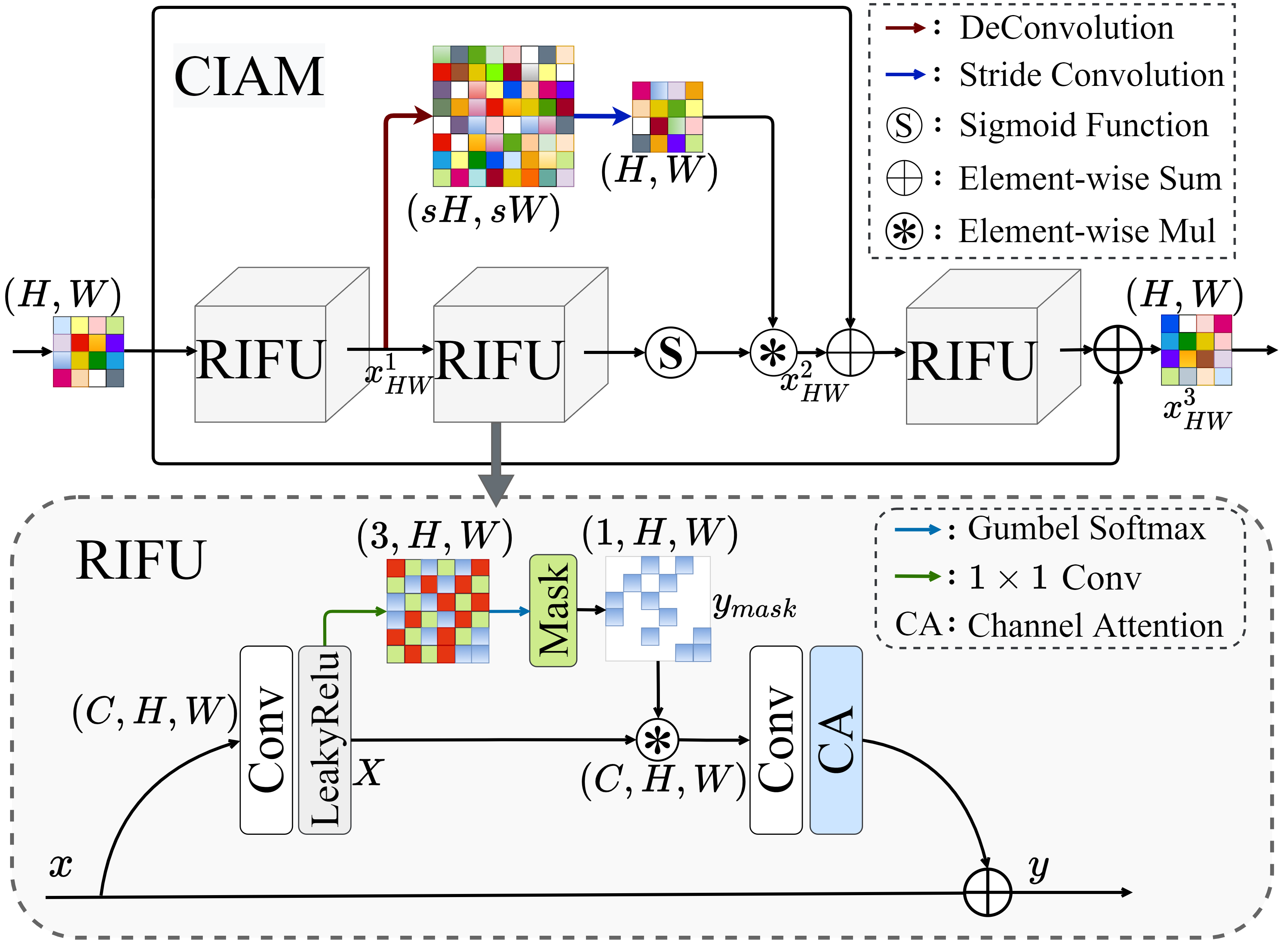}
\caption{The framework of Cross-scale Information Aggregation Module (CIAM) and Redundant Information Filter Unit (RIFU).}
\label{CIAM}
\end{figure}

According to previous work~\cite{zhang2019making}, we can know that it is easier to recover the smooth area that occupies most of the image area, but the complex texture information that occupies a small area of the image is difficult to recover. However, most SR methods tend to treat all areas of the image equally, which leads to the smooth area that accounts for most areas of the image being paid more attention by the network, which may miss the correct texture information, so accurate modeling cannot be achieved. In previous work~\cite{magid2021dynamic}, some methods convert the image from the time domain to the frequency domain by discrete cosine transform (DCT), and then manually set a threshold $T$ to discard the frequency domain information greater or less than $T$ to filter the secondary information. However, we also found that the input images used for training vary widely. 
Using manual thresholding is sensitive to noise, which is not suitable for all images. Therefore, we aim to explore a method that can make the network find a mask that instructs the model to pay more attention to texture features. To achieve this, we propose a  Redundant Information Filter Unit (RIFU). As given in Fig.~\ref{CIAM}, after the input feature $x$ enters RIFU, it first goes through a convolutional layer and an activation layer to obtain a mixed feature $X$. Then, we use a $1 \times 1$ convolutional layer to transform the number of channels of the output $R$ to $M$ ($M=3$), and its process can be formulated as
\begin{equation}
X = f_{lrelu}(f^{3x3}_{conv}(x)),
\end{equation}
\begin{equation}
R = f_{n -> 3}(X),
\end{equation}
where $f^{3x3}_{conv}(\cdot)$ represents the $3 \times 3$ convolutional layer, $f_{lrelu}{\cdot}$ represents the LeakyRelu function, and $f_{n -> 3}$ represents the $1 \times 1$ convolutional layer. Next, unlike channel attention or spatial attention, which often use pooling layers to accomplish redundant feature removal, we use the Gumbel Softmax trick~\cite{jang2016categorical} to generate a continuous differentiable normal distribution, which can well approximate the probability distribution represented by the network output and randomly add some sampling. Thus, the Gumbel-Softmax enables RIFU to retain some potentially useful information in addition to salient textural features. The process can be formulated as follows
\begin{equation}
G{S_i} = \frac{{\exp (({R_i} + g{s_i})/\tau )}}{{\sum\nolimits_{m = 1}^M {\exp (({R_{{i_m}}} + g{s_{{i_m}}})/\tau )} }},
\end{equation}
where $\tau $ defaults to 1, and ${g{s_i}}$ represents the noise obeying the $Gumbel(0,1)$ distribution. During training, we need the model to select only one channel from the three channels and the channel selection formula is as follows 
\begin{equation}
y_{mask} = \arg one(GS_{i_m}),
\end{equation}
where $y_{mask}$ represents the single-channel output feature after masking, and $\arg one(.)$ represents the argmax branch of $G{S_i}$ in $m$ channels. After that, we multiply $X$ by the $y_{mask}$ feature mask to preserve the initial details, then a convolutional layer and an attention mechanism are subsequently added to focus on the refined features and get the final output $y$ of RIFU through residual connection
\begin{equation}
y = f_{CA}(f^{3x3}_{conv}(y_{mask} \times X)) + x,
\end{equation}
where $f_{CA}(\cdot)$ represents the channel attention mechanism.

\subsubsection{Cross-scale Information Aggregation Module (CIAM)}
Due to the difference in model structure, the learned features are inevitably redundant. Meanwhile, with limited computing resources, we hope that the model will pay more attention to features with higher priority. So CIAM is designed to efficiently combine RIFU, which role is to extract more potentially effective information, mine deep image features, and make full use of information from different scales to observe image features.

In terms of network architecture design, simply combining RIFU in series is not conducive to gradient flow and collecting contextual information at different spatial locations. To solve this problem, we innovatively use two different scale-spaces for feature transformation in the middle of the module. Among them, one is the original space, whose feature map size has the same resolution as the input, and the other is the large space after the deconvolution operation. The receptive field of the transformed embedding is very large and can be used to guide the feature transformation in the original feature map. The process can be defined as
\begin{equation}
x^{1}_{HW} = f_{RIFU}(x_{HW}),
\end{equation}
\begin{equation}
x^{2}_{HW} = f_{sconv}(f_{deconv}(x^{1}_{HW})) \times f_{sig}(f_{RIFU}(x^{1}_{HW})),
\end{equation}
where $x^{i}_{HW}$ represents the output features with the size of $H \times W$ for each stage within the module, $f_{RIFU}(\cdot)$ represents the proposed RIFU, $f_{sconv}(\cdot)$ and $f_{deconv}(\cdot)$ represent the deconvolution and strided convolution with $s=2$, and $f_{sig}(\cdot)$ represents the Sigmoid function. 

Finally, the dense residual learning mechanism is introduced into the module to prevent the vanishing gradient. Therefore, this module has the potential for multiple RIFUs permutations and combinations, and its output $x^{3}_{HW}$ can be formulated as
\begin{equation}
x^{3}_{HW} = f_{RIFU}(x^{2}_{HW} + x_{HW}) + x_{HW}.
\end{equation}

\begin{figure}[t]
	\centerline{\includegraphics[width=8.5cm,trim=0 0 20 0]{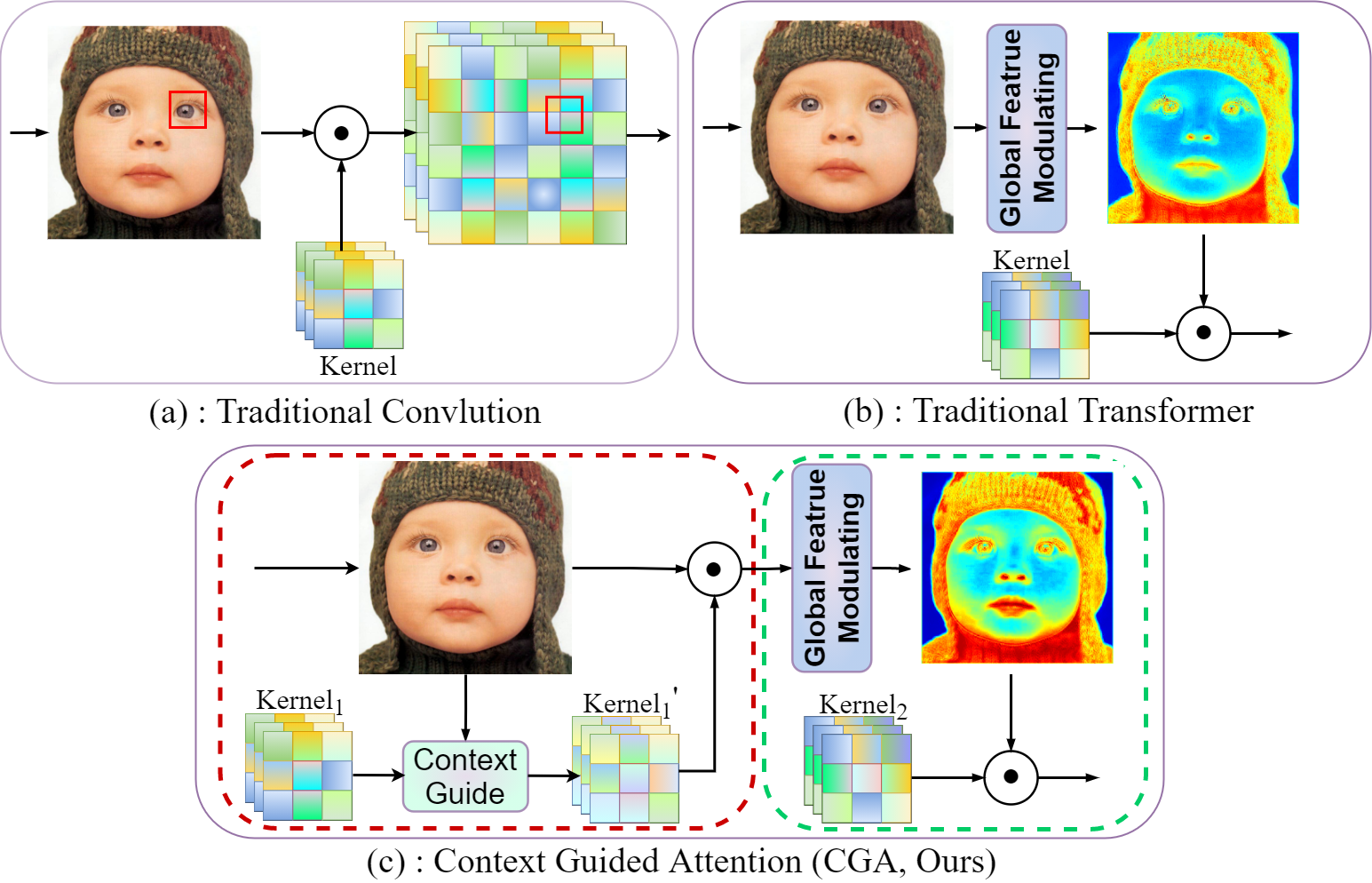}}
	\caption{(a) Traditional convolution focuses on local features. (b) Traditional Transformer uses global feature interactions to focus on key information. (c) Our proposed Context Guided Attention (CGA), guided by locally representative modulated convolution kernels, can adaptively combine context information to modify feature maps. Among them, the red box represents the Context Guided MaxConv (CGM) operation and the green box represents the self-attention, and $\odot$ denotes the convolution operation.}
	\label{CGA}
\end{figure}

\begin{figure*}
\begin{center}
\includegraphics[width=0.95\linewidth]{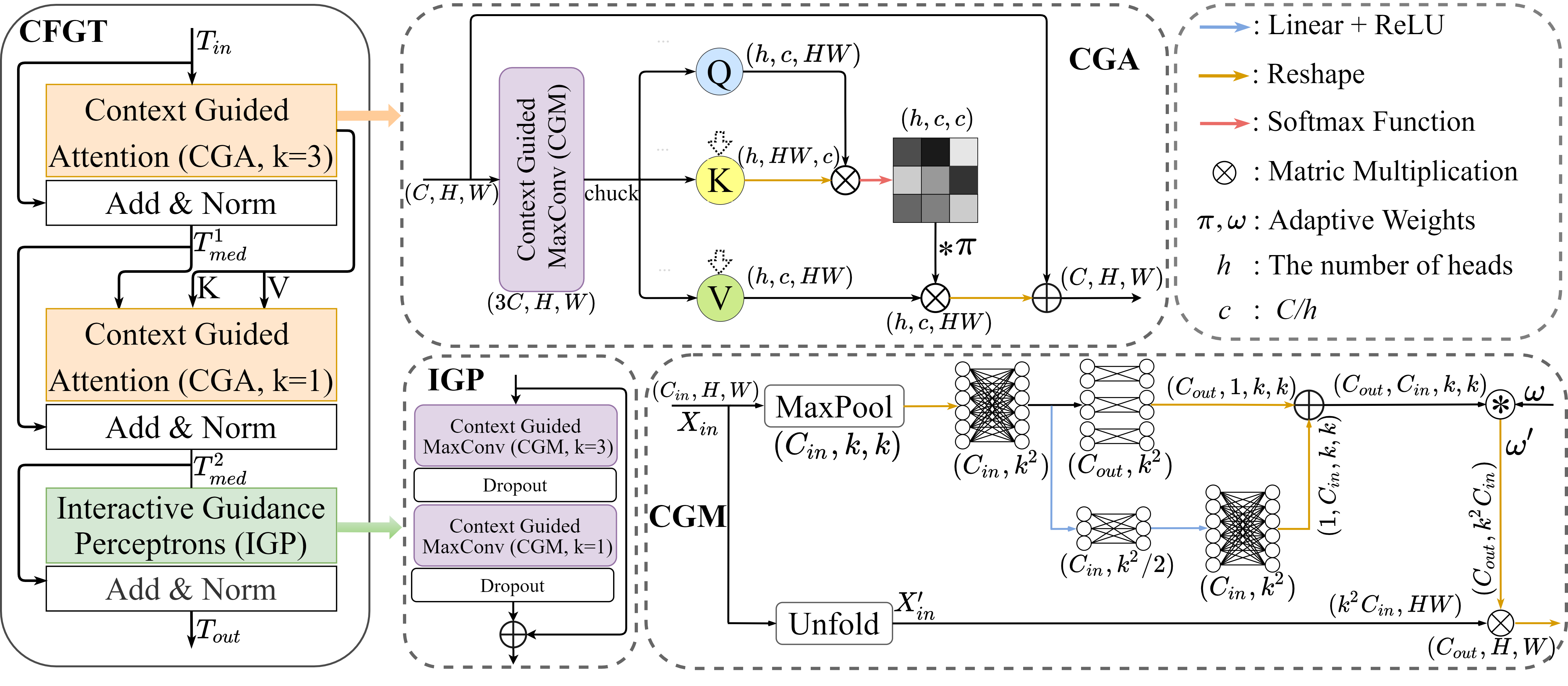}
\end{center}
   \caption{The architecture of the proposed Cross-receptive Field Guide Transformer (CFGT), Context Guided Attention (CGA), Context Guided MaxConv (CGM), and Interactive Guidance Perceptrons (IGP).}
\label{Transformer}
\end{figure*}

\subsection{Transformer Stage}
In recent years, Transformer has shown great potential on SISR, which can learn global information of images through its powerful self-attention mechanism. However, existing self-attention mechanisms often neglect to incorporate contextual features to build attention. To address this issue, we propose a Cross-receptive Field Guided Transformer (CFGT) in the Transformer stage. In CFGT, a specially designed Context Guided MaxConv (CGM) is introduced as its basic unit, which can adjust the weights of the network adaptively by reasoning the current semantic information, thus obtaining the features needed for the current pixel reconstruction from the contextual information. It is worth noting that semantic information usually refers to features of the image itself, such as texture, color, attributes, etc. It also meets the requirements of long-range modeling by establishing connections between the upper and lower CGAs across scalable perceptual fields.

\subsubsection{Context Guided Attention (CGA)}
As shown in Fig.~\ref{CGA}, traditional convolution focuses on local features, and traditional Transformer uses self-attention to capture the global Information. However, the traditional Transformer cannot take into account the locally dominant features, and the self-attention mechanism is also unable to incorporate contextual information. To alleviate this problem, we propose a new attention, named Context Guided Attention (CGA) (illustrated in Fig.~\ref{CGA} (c)), the red box represents the Context Guided MaxConv (CGM) module, where the modulated convolutional kernel uses the extracted semantic information to dynamically modulate its own weights to achieve an autonomous selection of information from the input context. Among them, the green box represents the self-attention mechanism, which performs global feature extraction based on the guided information output by the CGM in the red box. 

According to Fig.~\ref{Transformer}, we can clearly see that before computing the feature covariance to generate the global attention map, we introduce the Context Guided MaxConv (CGM) to emphasize the local context. Specifically, we first generate query ($Q$), key ($K$), and value ($V$) projections from the input tensor $X \in {\Re ^{C \times H \times W}}$. After that, we separately add representative local contexts to them, which are obtained by encoding the channel context through CGM $Q = W_{CGM}^Q(X)$, $K = W_{CGM}^K(X)$, and $V = W_{CGM}^V(X)$, $(Q,K,V) \in {\Re ^{h \times C/h \times HW}}$. Among them, $h$ is the amount of attention and $W_{CGM}(\cdot)$ represents the proposed CGM, similar to traditional multi-head attention~\cite{vaswani2017attention}. In this work, we divide the number of channels into $h$ groups for parallel learning, and the size of $Q, K, V$ is obtained after reshaping the tensor from the input image. Finally, we take the dot-product to reshape the $Q$ and $V$ projections to generate a transposed attention map of size ${\Re ^{h \times C/h \times C/h}}$ instead of a regular feature map of size ${\Re ^{h \times HW \times HW}}$~\cite{dosovitskiy2020image}. Overall, the attention mechanism in the first CGA can be formulated as 
\begin{equation}
\mathop {Att}\limits_{CG{A_1}} (Q,K,V) = V \cdot Soft((K \cdot Q)/\omega ),
\end{equation}
where $Soft(\cdot)$ represents the Softmax function, $ \cdot $ represents the dot product operation, $\omega$ represents a learnable adaptive parameter. Due to the powerful function of CGM combined with local features to explicitly modify the modulation kernel, the projection vector generated in the previous space can be used to guide the generation of subsequent attention.

\textbf{Context Guided MaxConv (CGM).}
In the convolution operation on the input feature ${X_{in}} \in {\Re ^{{C_{in}} \times H \times W}}$, a local feature block of size $k\times k$ ($k$ is the size of the convolution kernel) is extracted through a sliding window, and the extracted feature block is subsequently multiplied by the convolution kernel. Such a convolution operation can only extract local features and cannot adaptively affect the kernel based on the current semantic information. Recently, methods for context reasoning~\cite{chen2018iterative, li2019visual} have been extensively studied. Motivated by~\cite{lin2020context}, we propose a Context Guided MaxConv (CGM) to dynamically extract representative local patterns within Transformer in conjunction with contextual guidance. 

As shown in Fig.~\ref{Transformer}, to extract the representative information, we scale the input image to the size of $k \times k$ by using the max-pooling operation. Then, to alleviate the time-consuming kernel modulation caused by a large number of channels, we follow the idea of matrix decomposition~\cite{wei2020component} and reduce the complexity by generating two tensors through two branches. One of the branches draws on the idea of bottleneck design~\cite{hu2018squeeze}. We project the spatial position information into a vector of size ${{{k^2}} \mathord{\left/
 {\vphantom {{{k^2}} 2}} \right.
 \kern-\nulldelimiterspace} 2}$ through a linear layer of shared parameters and then generate new channel weights from this vector. In another branch, the idea of grouped convolution is applied to the linear layer, and the output dimension $C_{out}$ is obtained by introducing a grouped linear layer with weight. Then, reshape the two branches to obtain tensors with the size of ${C_{out}} \times 1 \times k \times k$ and $1 \times {C_{in}} \times k \times k$. Meanwhile, the two tensors are then summed by element to obtain our modulated convolution kernel, whose size is modulated to ${C_{in}} \times {C_{out}} \times k \times k$ to simulate the convolution kernel under real convolution operations. Subsequently, the simulated modulated convolution kernel is multiplied with the adaptive multiplier $\omega $. This process can be formulated as
\begin{equation}
\omega ' = kernel \times \omega  (\omega ' \in {\Re ^{{C_{in}} \times {C_{out}} \times k \times k}}) ,
\end{equation}
where $kernel$ represents our modulated convolutional kernel, and $\omega ' $ represents the adaptive modulated convolutional kernel.  It's worth noting that $\omega $ is an adaptive multiplier that is consistent with the size of the tensor. After multiplying it with the tensor, the tensor can be converted into a set of trainable type parameters and bound to the module. During the learning process, the weights of $\omega $ can be automatically learned and modified to optimize the model. On the other hand, the input ${X_{in}} \in {\Re ^{{C_{{\rm{in}}}} \times H \times W}}$ uses the unfold function to locally connect the sliding features of size $k \times k$ extracted from each sliding window to make contextual connections between different pixel features to obtain a feature map of size ${X_{in}}' \in {\Re ^{{k^2}{C_{in}} \times HW}}$. Finally, a modulated convolutional kernel $\omega '$ that fully understands the semantic information of the current key pixel can use the input ${X_{in}}'$ to obtain guidance on context to adaptively capture the contextual information needed for the key pixel.

\subsubsection{Cross-receptive Field Guide Transformer (CFGT)}
In this paper, we propose a Cross-receptive Field Guide Transformer (CFGT) for long-distance modeling. In contrast to previous Transformer-based methods, such as SwinIR~\cite{liang2021swinir} and LBNet~\cite{gao2022lightweight}, our CFGT shares the information within different self-attention for better global modeling. As shown in Fig.~\ref{Transformer}, CFGT is mainly composed of two CGAs and one Interactive Guidance Perceptron (IGP) in the encoder part. Meanwhile, hierarchical normalization is performed after each block, and the local residual connection is used. 
It is worth noting that the CGM in the two CGAs adopts different $k$ ($k$ represents the size of the receptive field that the CGM focuses on). Since a larger receptive field may bring more computational load, we simulate the common convolutional kernel size and chose $3 \times 3$ and $1 \times 1$ receptive fields. 
Therefore, cross-attention can supplement the model with cross-scale features. And the understanding of current pixel-level features to other pixel features is facilitated by communicating features across scales, thus allowing the model to acquire richer contextual information. We assume that the input of CFGT is $T_{in}$, then the output ${T_{out}}$ can be formulated as
\begin{equation}
{T^{1}_{med}} = Norm(CGA({T_{in}})) + {T_{in}},
\end{equation}
\begin{equation}
{T^{2}_{med}} = Norm(CGA({T^{1}_{med}},K,V)) + {T^{1}_{med}},
\end{equation}
\begin{equation}
{T_{out}} = Norm(IGP({T^{2}_{med}})) + {T^{2}_{med}},
\end{equation}
where $K$ and $V$ are the key and value generated by the first CGA, \textbf{and they will serve as part of the input of the second CGA and interact with the $Q'$, $K'$, $V'$ generated by the second CGA.} The operation of the second CGA can be defined as
\begin{small}
\begin{equation}
\mathop {Att}\limits_{CG{A_2}} (Q',K',V',K,V) = (V' + V) \cdot Soft(((K' + K) \cdot Q')/\omega ),
\end{equation}
\end{small}

Similarly, our IGP also adopts the idea of cross-receptive fields, connecting a CGM with $k=3$ and a CGM with $k=1$ using a residual structure to enhance the attention of the Transformer. In general, contextual interaction within CFGT mainly consists of two main parts. One is the interaction brought by capturing the contextual information contained in the input through the modulated convolutional kernel within the CGM. The other is the interaction between the upper and lower two CGAs through the communication of vectors $K$ and $V$ with different sizes of receptive fields.

\subsection{Loss Function}
During training, given a training set $\left\{ {I_i^{LR},I_i^{HR}} \right\}_{i = 1}^N$, the loss function of CFIN can be expressed by
\begin{equation}
Loss\left( \theta  \right) = \arg \mathop {\min }\limits_\theta  \frac{1}{N}{\sum\limits_{i = 1}^N {\left\| {{F_{CFIN}}(I_{LR}^i) - I_{HR}^i} \right\|} _1},
\end{equation}
where ${{F_{CFIN}}}(\cdot)$ represent our proposed CFIN, $\theta $ represents the parameter set of CFIN, and $N$ represents the number of LR-HR pairs in the training set.

\section{Experiments}~\label{EX}
In this part, we provide relevant experimental details, descriptions, and results to verify the effectiveness and excellence of the proposed CFIN.

\begin{table*}[t]
    \centering
    \small
    \caption{Performance comparisons with other advanced CNN-based SISR models. The best and the second-best results are highlighted and underlined, respectively. '+' indicates that the model uses the self-ensemble strategy, which is to average the results of the original image, the original horizontal flip, the original vertical flip, and the original vertical flip.}
    \scalebox{0.9}{
    \begin{adjustbox}{width=1\linewidth}
    \begin{tabular}{|l|c|c|c|c|c|c|c|c|}
    \hline
    \multirow{2}{*}{Methods} & \multicolumn{1}{l|}{\multirow{2}{*}{Scale}} & \multirow{2}{*}{Params}& \multirow{2}{*}{Multi-adds} & Set5 & Set14 & BSDS100 & Urban100 & Manga109 \\
    \cline{5-9} 
    & \multicolumn{1}{l|}{} & &  & PSNR/SSIM & PSNR/SSIM & PSNR/SSIM & PSNR/SSIM & PSNR/SSIM \\
    \hline
    \hline
    
    IDN~\cite{hui2018fast}  & \multirow{18}{*}{$\times 2$}       & 553K    &124.6G    & 37.83/0.9600       & 33.30/0.9148             &  32.08/0.8985    & 31.27/0.9196     & 38.01/0.9749 \\
    CARN~\cite{ahn2018fast}                        &        & 1592K  &222.8G   & 37.76/0.9590       & 33.52/0.9166           & 32.09/0.8978    & 31.92/0.9256     & 38.36/0.9765 \\
    IMDN~\cite{hui2019lightweight}                &       & 694K    &158.8G   & 38.00/0.9605       & 33.63/0.9177           &  32.19/0.8996    & 32.17/0.9283     & 38.88/0.9774 \\
    AWSRN-M~\cite{wang2019lightweight}     &       & 1063K  &244.1G   & 38.04/0.9605       & 33.66/0.9181           &  32.21/ 0.9000    & 32.23/0.9294    & 38.66/0.9772 \\
    MADNet~\cite{lan2020madnet}                 &        & 878K    &187.1G   & 37.85/0.9600        & 33.38/0.9161            & 32.04/0.8979    & 31.62/0.9233     & - \\
    MAFFSRN-L~\cite{muqeet2020multi}       &       & 790K    &154.4G   & 38.07/0.9607       & 33.59/0.9177           & 32.23/0.9005    & 32.38/0.9308     & - \\
    LAPAR-A~\cite{li2020lapar}                     &       & 548K    &171.0G   & 38.01/0.9605        & 33.62/0.9183            & 32.19/0.8999    & 32.10/0.9283   &38.67/0.9772  \\
    RFDN~\cite{liu2020residual}      &       	& 534K  &123.0G  & 38.05/0.9606        &33.68/0.9184            & 32.16/0.8994    & 32.12/0.9278     & 38.88/0.9773 \\
    LatticeNet+~\cite{luo2020latticenet}           &       & 756K    &165.5G   & \underline{38.15}/0.9610        & 33.78/0.9193            & 32.25/0.9005    & 32.43/0.9302     & - \\
    SMSR~\cite{wang2021exploring}                &       & 985K    &351.5G   & 38.00/0.9601       &33.64/0.9179            & 32.17/0.8990    & 32.19/0.9284     &38.76/0.9771 \\
    PFFN~\cite{zhang2021pffn}         		& 	 & 569K	& 138.3G	  & 38.07/0.9607       & 33.69/0.9192             & 32.21/0.8997   & 32.33/0.9298   &38.89/\underline{0.9775}\\
    DRSAN~\cite{park2021dynamic}                     &       & 690K    &159.3G   & 38.11/0.9609        & 33.64/0.9185            & 32.21/0.9005    & 32.35/0.9304   & -  \\
    FDIWN~\cite{gao2022feature}         		& 	 & 629K	& 112.0G	  & 38.07/0.9608       & 33.75/\underline{0.9201}            & 32.23/0.9003   & 32.40/0.9305   & 38.85/0.9774 \\
    LatticeNet-CL~\cite{luo2022lattice}           &       & 756K    &169.5G   & 38.09/0.9608        & 33.70/0.9188            & 32.21/0.9000    & 32.29/0.9291     & - \\
    FMEN~\cite{du2022fast}         		& 	 & 748K	& 172.0G	  & 38.10/0.9609       & 33.75/0.9192           & 32.26/0.9003   & 32.41/0.9311   & 38.95/0.9778 \\
    \textbf{CFIN (Ours)}                               &      &675K     &116.9G  & 38.14/ \underline{0.9610}   & \underline{33.80}/0.9199        & \underline{32.26}/ \underline{0.9006}    &\underline{32.48}/ \underline{0.9311}   &\underline{38.97}/0.9773     \\
    \textbf{CFIN+ (Ours)}                           &       &675K    &116.9G  & \textbf{38.22}/ \textbf{0.9613}   & \textbf{34.01}/ \textbf{0.9221}        & \textbf{32.35}/ \textbf{0.9016}    &\textbf{32.93}/ \textbf{0.9347}   &\textbf{39.21}/ \textbf{0.9777}     \\

    \hline
    \hline
    
    IDN~\cite{hui2018fast}      & \multirow{18}{*}{$\times 3$}        & 553K    &56.3G  & 34.11/0.9253        & 29.99/0.8354         & 28.95/0.8013    & 27.42/0.8359   & 32.71/0.9381   \\   
    CARN~\cite{ahn2018fast}                           &        & 1592K   &118.8G & 34.29/0.9255        &30.29/0.8407            & 29.06/0.8034    & 28.06/0.8493   & 33.43/0.9427    \\
    IMDN~\cite{hui2019lightweight}                 &        & 703K    &71.5G  & 34.36/0.9270       & 30.32/0.8417           & 29.09/0.8046    & 28.17/0.8519      & 33.61/0.9445 \\
    AWSRN-M~\cite{wang2019lightweight}      &        & 1143K  &116.6G  & 34.42/0.9275       & 30.32/0.8419           & 29.13/0.8059    & 28.26/0.8545    & 33.64/0.9450 \\
    MADNet~\cite{lan2020madnet}                  &       & 930K     &88.4G  & 34.16/0.9253        & 30.21/0.8398           & 28.98/0.8023    & 27.77/0.8439           & -    \\
    MAFFSRN-L~\cite{muqeet2020multi}       &       & 807K    &68.5G & 34.45/0.9277        & 30.40/0.8432            & 29.13/0.8061    & 28.26/0.8552     & - \\
    LAPAR-A~\cite{li2020lapar}                      &       & 594K    &114.0G & 34.36/0.9267        & 30.34/0.8421            &29.11/0.8054    & 28.15/0.8523   & 33.51/0.9441   \\
    RFDN~\cite{liu2020residual}       &        & 541K  &55.4G  & 34.41/0.9273        & 30.34/0.8420            & 29.09/0.8050    & 28.21/0.8525     & 33.67/0.9449 \\
    LatticeNet+~\cite{luo2020latticenet}               &       & 765K    &76.3G & 34.53/0.9281        & 30.39/0.8424           &29.15/0.8059    & 28.33/0.8538   & -   \\
    SMSR~\cite{wang2021exploring}                &       & 993K    &156.8G& 34.40/0.9270       &30.33/0.8412           & 29.10/0.8050    & 28.25/0.8536     & 33.68/0.9445\\
    PFFN~\cite{zhang2021pffn}                       &       & 558K    &69.1G & 34.54/0.9282        & 30.42/0.8435          &29.17/0.8062    & 28.37/0.8566   & 33.63/0.9455   \\
    DRSAN~\cite{park2021dynamic}                     &       & 740K    &76.0G   & 34.50/0.9278        & 30.39/0.8437            & 29.13/0.8065    & 28.35/0.8566   & -  \\
    FDIWN~\cite{gao2022feature}                   &        & 645K    &51.5G  &34.52/0.9281         &30.42/0.8438          &29.14/0.8065   &28.36/0.8567     & 33.77/0.9456 \\
    LatticeNet-CL~\cite{luo2022lattice}           &       & 765K    &76.3G   & 34.46/0.9275        & 30.37/0.8422            & 29.12/0.8054    & 28.23/0.8525     & - \\
    FMEN~\cite{du2022fast}                   &        & 757K    &77.2G  &34.45/0.9275         &30.40/0.8435          &29.17/0.8063   &28.33/0.8562     & 33.86/0.9462 \\
    \textbf{CFIN (Ours)}                              &        &681K     &53.5G  &  \underline{34.65}/ \underline{0.9289}  & \underline{30.45}/ \underline{0.8443}   & \underline{29.18}/ \underline{0.8071}    & \underline{28.49}/ \underline{0.8583}    & \underline{33.89}/ \underline{0.9464}  \\
    \textbf{CFIN+ (Ours)}                            &        &681K      &53.5G  & \textbf{34.75}/\textbf{0.9298}  &\textbf{30.59}/\textbf{0.8467}   &\textbf{29.27}/\textbf{0.8091}    &\textbf{28.85}/\textbf{0.8645}    &\textbf{34.26}/\textbf{0.9484}  \\

    \hline
    \hline
    
    IDN~\cite{hui2018fast}  & \multirow{19}{*}{$\times 4$}       & 553K   &32.3G & 31.82/0.8903        & 28.25/0.7730         & 27.41/0.7297       & 25.41/0.7632   &29.41/0.8942     \\ 
    CARN~\cite{ahn2018fast}                          &        & 1592K  & 90.9G & 32.13/0.8937        & 28.60/0.7806            & 27.58/0.7349    & 26.07/0.7837   & 30.42/0.9070    \\
    IMDN~\cite{hui2019lightweight}                 &        & 715K   &40.9G  & 32.21/0.8948       & 28.58/0.7811          & 27.56/0.7353      & 26.04/0.7838      &30.45/0.9075 \\
    AWSRN-M~\cite{wang2019lightweight}     &        & 1254K  &72.0G  & 32.21/0.8954    & 28.65/0.7832           & 27.60/0.7368      & 26.15/0.7884     & 30.56/0.9093 \\
    MADNet~\cite{lan2020madnet}                 &         & 1002K &54.1G & 31.95/0.8917       & 28.44/0.7780            & 27.47/0.7327    & 25.76/0.7746   & -    \\
    MAFFSRN-L~\cite{muqeet2020multi}       &        & 830K   &38.6G & 32.20/0.8953       & 28.62/0.7822            & 27.59/0.7370    & 26.16/0.7887    & - \\
    LAPAR-A~\cite{li2020lapar}                    &        & 659K   &94.0G  &32.15/0.8944        &28.61/0.7818            &27.61/0.7366    & 26.14/0.7871   &30.42/0.9074    \\
    RFDN~\cite{liu2020residual}       &        & 550K  &31.6G  & 32.24/0.8952        & 28.61/0.7819            & 27.57/0.7360    & 26.11/0.7858     & 30.58/0.9089 \\
    LatticeNet+~\cite{luo2020latticenet}             &        & 777K  &43.6G  &32.30/0.8962        &28.68/0.7830          &27.62/0.7367    & 26.25/0.7873        &-    \\
    SMSR~\cite{wang2021exploring}                &       & 1006K&89.1G& 32.12/0.8932          &28.55/0.7808           &27.55/0.7351    &26.11/0.7868     & 30.54/0.9085\\
    PFFN~\cite{zhang2021pffn}                     &        & 569K  &45.1G  &32.36/0.8967        &28.68/0.7827            &27.63/0.7370    & 26.26/0.7904   &30.50/0.9100    \\
    DRSAN~\cite{park2021dynamic}                     &       & 730K    &49.0G   & 32.30/0.8954        & 28.66/0.7838            & 27.61/0.7381    & 26.26/0.7920   & -  \\
    FDIWN~\cite{gao2022feature}                &       & 664K  &28.4G   & 32.23/0.8955   &28.66/0.7829            &27.62/0.7380        &26.28/0.7919                 & 30.63/0.9098 \\
    LatticeNet-CL~\cite{luo2022lattice}          &       & 777K  &43.6G   & 32.30/0.8958   &28.65/0.7822            &27.59/0.7365        &26.19/0.7855                 & - \\
    FMEN~\cite{du2022fast}                &       & 769K  & 44.2G   & 32.24/0.8955   &28.70/0.7839            &27.63/0.7379        &26.28/0.7908                 & 30.70/0.9107 \\
    \textbf{CFIN (Ours)}                            &        &699K  &31.2G  & \underline{32.49}/\underline{0.8985}    & \underline{28.74}/\underline{0.7849}    & \underline{27.68}/\underline{0.7396}   &  \underline{26.39}/\underline{0.7946}       & \underline{30.73}/\underline{0.9124}     \\
    \textbf{CFIN+ (Ours)}                          &        &699K    &31.2G  &\textbf{32.60}/\textbf{0.8998}    &\textbf{28.86}/\textbf{0.7871}     & \textbf{27.76}/\textbf{0.7419}    & \textbf{26.71}/\textbf{0.8028}       &\textbf{31.15}/\textbf{0.9163}     \\
    \hline
    \end{tabular}
    \end{adjustbox}
    }
    \label{tab:sota}
\end{table*}

\subsection{Datasets and Metrics}
Following previous works, we utilize DIV2K~\cite{timofte2017ntire} (1-800) as our training dataset. Meanwhile, we used five benchmark datasets to verify the effectiveness of the proposed model, including Set5~\cite{bevilacqua2012low}, Set14~\cite{zeyde2010single}, BSDS100~\cite{martin2001database}, Urban100~\cite{huang2015single}, and Manga109~\cite{matsui2017sketch}. Additionally, we used Peak Signal-to-Noise Ratio (PSNR) and Structural Similarity (SSIM) to evaluate the quality of our restored images on the Y channel of the YCbCr color space.

\subsection{Implementation Details}
During training, we randomly crop patches with the size of $48 \times 48$ from the training set as input and use horizontal flipping and random rotation for data augmentation. The initial learning rate is set to $5 \times {10^{ - 4}}$, which is finally reduced to ${\rm{6}}{\rm{.25}} \times {\rm{1}}{{\rm{0}}^{ - 6}}$ by cosine annealing. Meanwhile, we implement our model with the PyTorch framework and update it with Adam optimizer. All our experiments are conducted on NVIDIA RTX 2080Ti GPU. In the final model CFIN, we use 8 transformer stages and 8 convolution stages, all of which are called twice using a loop mechanism. Meanwhile, we set the initial input channel to 48 and use the weight normalization~\cite{salimans2016weight} after each convolutional layer in RIFUs.

\begin{table*}
	\centering
	\small
	\caption{Comparisons with some Transformer-base methods for $\times 4$ SR. * means this model is pre-trained based on the $\times 2$ setup and the training patch size is 64 $\times $ 64 (ours is 48 $\times $ 48 and without pre-training). CFIN-L denotes a larger version.}
	\label{tab:swinir}
	\scalebox{0.8}{
	\begin{tabular}{|c|c|c|c|c|c|c|c|c|c|c|}
		\hline
		\multirow{2}{*}{Methods} & \multirow{2}{*}{Params} & \multirow{2}{*}{Multi-adds} & \multirow{2}{*}{GPU}  & \multirow{2}{*}{Time}  & Set5 & Set14 & BSD100 & Urban100 & Manga109 & Average  \\
		\cline{6-11}
		& & & & &PSNR / SSIM & PSNR / SSIM & PSNR / SSIM & PSNR / SSIM & PSNR / SSIM & PSNR / SSIM \\
		\hline
		\hline
		SwinIR*~\cite{liang2021swinir} & 897K & 49.6G &10500M & 0.046s  & 32.44 / 0.8976 & \textbf{28.77} / \textbf{0.7858} &  \textbf{27.69 / 0.7406} & \underline{26.47} / \textbf{0.7980} & \textbf{30.92} / \textbf{0.9151} & \underline{29.26} / \textbf{0.8274}\\
		ESRT~\cite{lu2021efficient} & 751K & 67.7G &4191M &0.032s   & 32.19 / 0.8947 & 28.69 / 0.7833 &  \textbf{27.69} / 0.7379 & 26.39 / 0.7962 & 30.75 / 0.9100 & 29.14 / 0.8244\\
		LBNet~\cite{gao2022lightweight} & 742K & 38.9G & 6417M & 0.043s  & 32.29 / 0.8960 & 28.68 / 0.7832 &  27.62 / 0.7382 & 26.27 / 0.7906 & 30.76 / 0.9111 & 29.12 / 0.8238\\
		CFIN & 699K & 31.2G &11453M & 0.035s  & \underline{32.49 / 0.8985} & \underline{28.74} / 0.7849 &  \underline{27.68 / 0.7396} & 26.39 / 0.7946 & 30.73 / 0.9124 & 29.21 / 0.8260\\
		CFIN-L & 852K & 37.8G &14585M & 0.040s  & \textbf{32.56} / \textbf{0.8988} & \underline{28.74 / 0.7852} &  \textbf{27.69} / \textbf{0.7406} & \textbf{26.49} / \underline{0.7973} & \underline{30.85 / 0.9134} & \textbf{29.27} / \underline{0.8271}\\
		\hline
	\end{tabular}
	}
\end{table*}

\begin{figure*}[htpb]
	\scriptsize
	\centering
	\scalebox{0.88}{
		\begin{tabular}{lc}
            \begin{adjustbox}{valign=t}
				\begin{tabular}{c}				    
                    \includegraphics[width=0.25\textwidth, height=0.145\textheight]{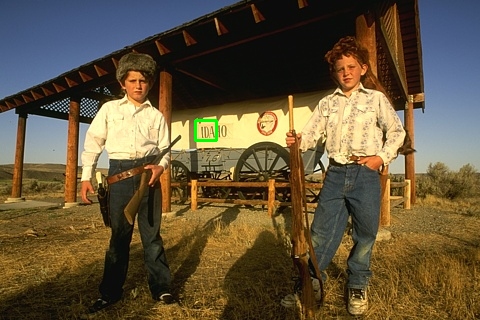} \\
					BSDS100 ($\times 2$): \\
					21681 \\
				\end{tabular}
			\end{adjustbox}
			\hspace{-3mm}
			\begin{adjustbox}{valign=t}
				\begin{tabular}{ccccc}
                    \includegraphics[width=0.15\textwidth, height=0.06\textheight]{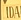} & 
					\hspace{-3mm}
					\includegraphics[width=0.15\textwidth, height=0.06\textheight]{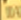} & 
					\hspace{-3mm}
					\includegraphics[width=0.15\textwidth, height=0.06\textheight]{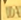} & 
                    \hspace{-3mm}
                    \includegraphics[width=0.15\textwidth, height=0.06\textheight]{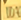} & 
                    \hspace{-3mm}
                    \includegraphics[width=0.15\textwidth, height=0.06\textheight]{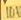}\\
					HR & \hspace{-2mm}
				  Bicubic & \hspace{-2mm}
					VDSR~\cite{kim2016accurate} & \hspace{-2mm}
                    CARN-M~\cite{ahn2018fast} & \hspace{-2mm}
                    CARN~\cite{ahn2018fast} \\
					PSNR/SSIM & \hspace{-2mm}
					19.15/0.6690 & \hspace{-2mm}
					22.37/0.8185 & \hspace{-2mm}
                    22.54/0.8261 & \hspace{-2mm}
					23.72/0.8511 \\
                    \includegraphics[width=0.15\textwidth, height=0.06\textheight]{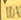} & 
                    \hspace{-3mm}
					\includegraphics[width=0.15\textwidth, height=0.06\textheight]{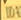} & 
					\hspace{-3mm}
					\includegraphics[width=0.15\textwidth, height=0.06\textheight]{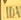} & 
					\hspace{-3mm}
                    \includegraphics[width=0.15\textwidth, height=0.06\textheight]{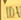} & 
					\hspace{-3mm}
					\includegraphics[width=0.15\textwidth, height=0.06\textheight]{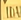} \\
                    IMDN~\cite{hui2019lightweight} & \hspace{-2mm}
					MADNet~\cite{lan2020madnet} & \hspace{-2mm}
				  AWSRN-M~\cite{wang2019lightweight} & \hspace{-2mm}
                    FDIWN~\cite{gao2022feature} & \hspace{-2mm}
					\textbf{Ours} \\
					22.89/0.8337 & \hspace{-2mm}
					23.70/0.8510 & \hspace{-2mm}
                    24.05/0.8552 & \hspace{-2mm}
					23.84/0.8532& \hspace{-2mm}
					\textbf{24.40/0.8625} \\
				\end{tabular}
			\end{adjustbox}
             \\

             \begin{adjustbox}{valign=t}
				\begin{tabular}{c}
					\includegraphics[width=0.25\textwidth, height=0.145\textheight]{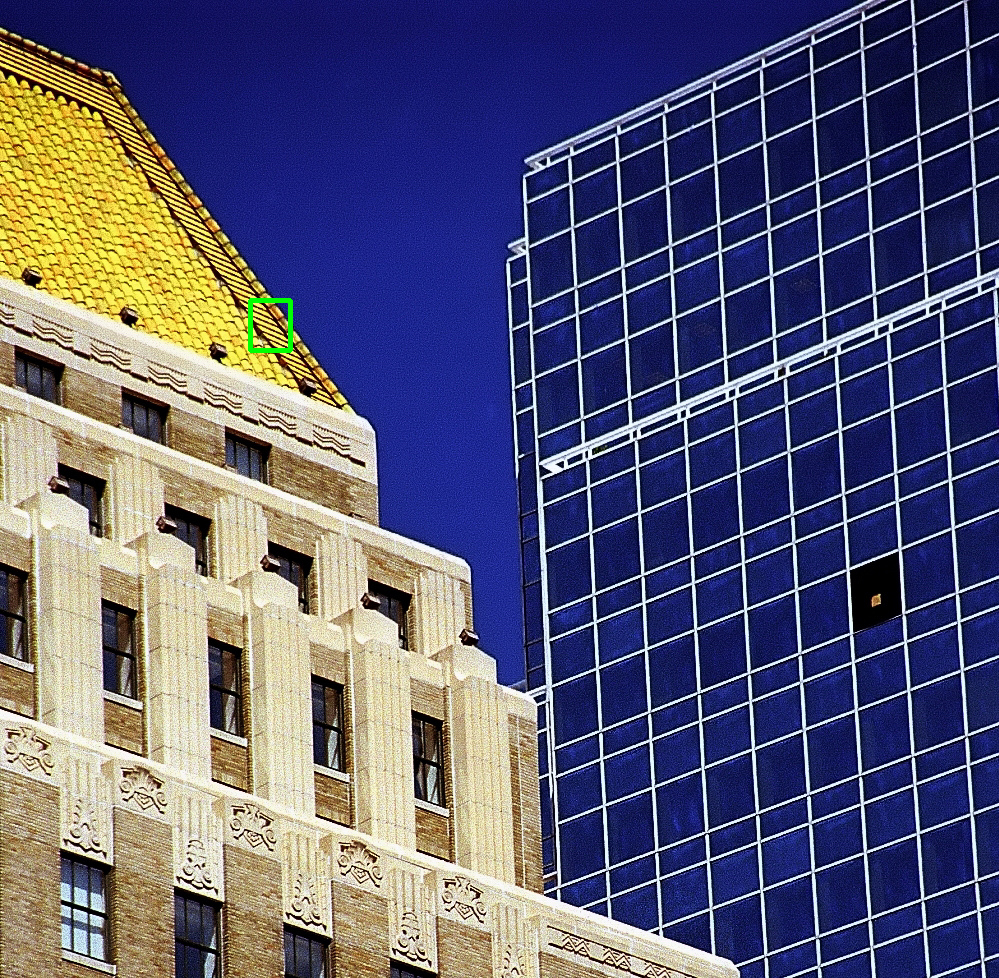} \\
					Urban100 ($\times 3$): \\
					img063 \\
				\end{tabular}
			\end{adjustbox}
			\hspace{-3mm}
			\begin{adjustbox}{valign=t}
				\begin{tabular}{ccccc}
					\includegraphics[width=0.15\textwidth, height=0.06\textheight]{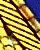} & 
					\hspace{-3mm}
					\includegraphics[width=0.15\textwidth, height=0.06\textheight]{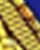} & 
					\hspace{-3mm}
					\includegraphics[width=0.15\textwidth, height=0.06\textheight]{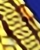} & 
                    \hspace{-3mm}
                    \includegraphics[width=0.15\textwidth, height=0.06\textheight]{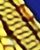} & 
                    \hspace{-3mm}
                    \includegraphics[width=0.15\textwidth, height=0.06\textheight]{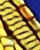}\\
					HR & \hspace{-2mm}
				  Bicubic & \hspace{-2mm}
					VDSR~\cite{kim2016accurate} & \hspace{-2mm}
                    CARN-M~\cite{ahn2018fast} & \hspace{-2mm}
                    CARN~\cite{ahn2018fast}\\
					PSNR/SSIM & \hspace{-2mm}
					19.66/0.5162 & \hspace{-2mm}
					23.14/0.6546 & \hspace{-2mm}
                    23.56/0.6651 & \hspace{-2mm}
					23.80/0.6715 \\
                    \includegraphics[width=0.15\textwidth, height=0.06\textheight]{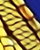} & 
                    \hspace{-3mm}
					\includegraphics[width=0.15\textwidth, height=0.06\textheight]{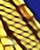} & 
					\hspace{-3mm}
					\includegraphics[width=0.15\textwidth, height=0.06\textheight]{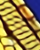} & 
					\hspace{-3mm}
                    \includegraphics[width=0.15\textwidth, height=0.06\textheight]{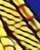} & 
					\hspace{-3mm}
					\includegraphics[width=0.15\textwidth, height=0.06\textheight]{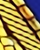} \\
                    IMDN~\cite{hui2019lightweight} & \hspace{-2mm}
					MADNet~\cite{lan2020madnet} & \hspace{-2mm}
				  AWSRN-M~\cite{wang2019lightweight} & 
                    \hspace{-2mm}
                    FDIWN~\cite{gao2022feature} & \hspace{-2mm}
					\textbf{Ours} \\
					23.88/0.6724 & \hspace{-2mm}
					24.03/0.6710 & \hspace{-2mm}
                    23.70/0.6736 & \hspace{-2mm}
                    24.33/0.6761 & \hspace{-2mm}
					\textbf{24.42/0.6767} \\
				\end{tabular}
			\end{adjustbox}
             \\

             \begin{adjustbox}{valign=t}
				\begin{tabular}{c}
					\includegraphics[width=0.25\textwidth, height=0.145\textheight]{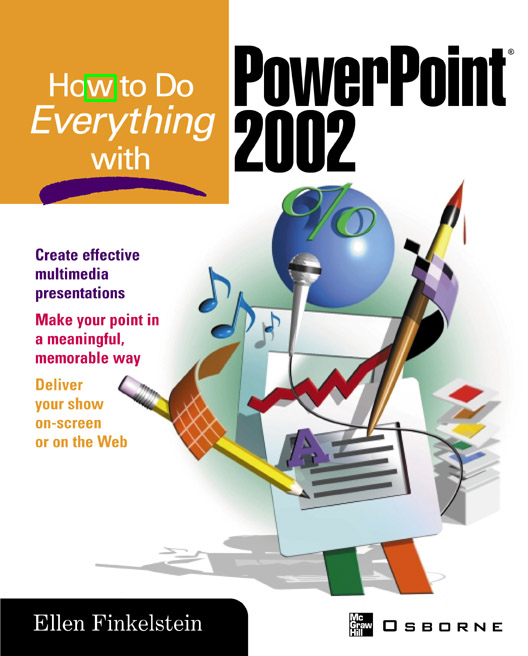} \\
					Set14 ($\times 4$): \\
					ppt \\
				\end{tabular}
			\end{adjustbox}
			\hspace{-3mm}
			\begin{adjustbox}{valign=t}
				\begin{tabular}{ccccc}
					\includegraphics[width=0.15\textwidth, height=0.06\textheight]{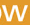} & 
					\hspace{-3mm}
					\includegraphics[width=0.15\textwidth, height=0.06\textheight]{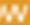} & 
					\hspace{-3mm}
					\includegraphics[width=0.15\textwidth, height=0.06\textheight]{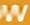} & 
                    \hspace{-3mm}
                    \includegraphics[width=0.15\textwidth, height=0.06\textheight]{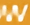} & 
                    \hspace{-3mm}
                    \includegraphics[width=0.15\textwidth, height=0.06\textheight]{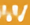}\\
					HR & \hspace{-2mm}
				  Bicubic & \hspace{-2mm}
					VDSR~\cite{kim2016accurate} & \hspace{-2mm}
                    CARN-M~\cite{ahn2018fast} & \hspace{-2mm}
                    CARN~\cite{ahn2018fast}\\
					PSNR/SSIM & \hspace{-2mm}
					22.21/0.8268 & \hspace{-2mm}
					26.02/0.9318 & \hspace{-2mm}
                    26.46/0.9430 & \hspace{-2mm}
					26.88/0.9505\\
                    \includegraphics[width=0.15\textwidth, height=0.06\textheight]{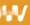} & 
                    \hspace{-3mm}
					\includegraphics[width=0.15\textwidth, height=0.06\textheight]{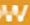} & 
					\hspace{-3mm}
					\includegraphics[width=0.15\textwidth, height=0.06\textheight]{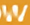} & 
					\hspace{-3mm}
                    \includegraphics[width=0.15\textwidth, height=0.06\textheight]{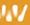} & 
					\hspace{-3mm}
					\includegraphics[width=0.15\textwidth, height=0.06\textheight]{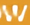} \\
                    IMDN~\cite{hui2019lightweight} & \hspace{-2mm}
					MADNet~\cite{lan2020madnet} & \hspace{-2mm}
				  AWSRN-M~\cite{wang2019lightweight} & 
                    \hspace{-2mm}
                    FDIWN~\cite{gao2022feature} & \hspace{-2mm}
					\textbf{Ours} \\
					26.74/0.9493 & \hspace{-2mm}
					26.38/0.9415 & \hspace{-2mm}
                    26.82/0.9512 & \hspace{-2mm}
                    26.75/0.9496 & \hspace{-2mm}
					\textbf{27.02/0.9531} \\
				\end{tabular}
			\end{adjustbox}
             \\

	\end{tabular} }
	\caption{Visual comparisons with different lightweight SISR models. Obviously, our CFIN can reconstruct high-quality images.}
	\label{visual}
\end{figure*}

\begin{figure*}[!t]
	\scriptsize
	\centering
	\scalebox{0.9}{
		\begin{tabular}{lc}
            \begin{adjustbox}{valign=t}
				\begin{tabular}{c}
				\includegraphics[width=0.2\textwidth, height=0.125\textheight]{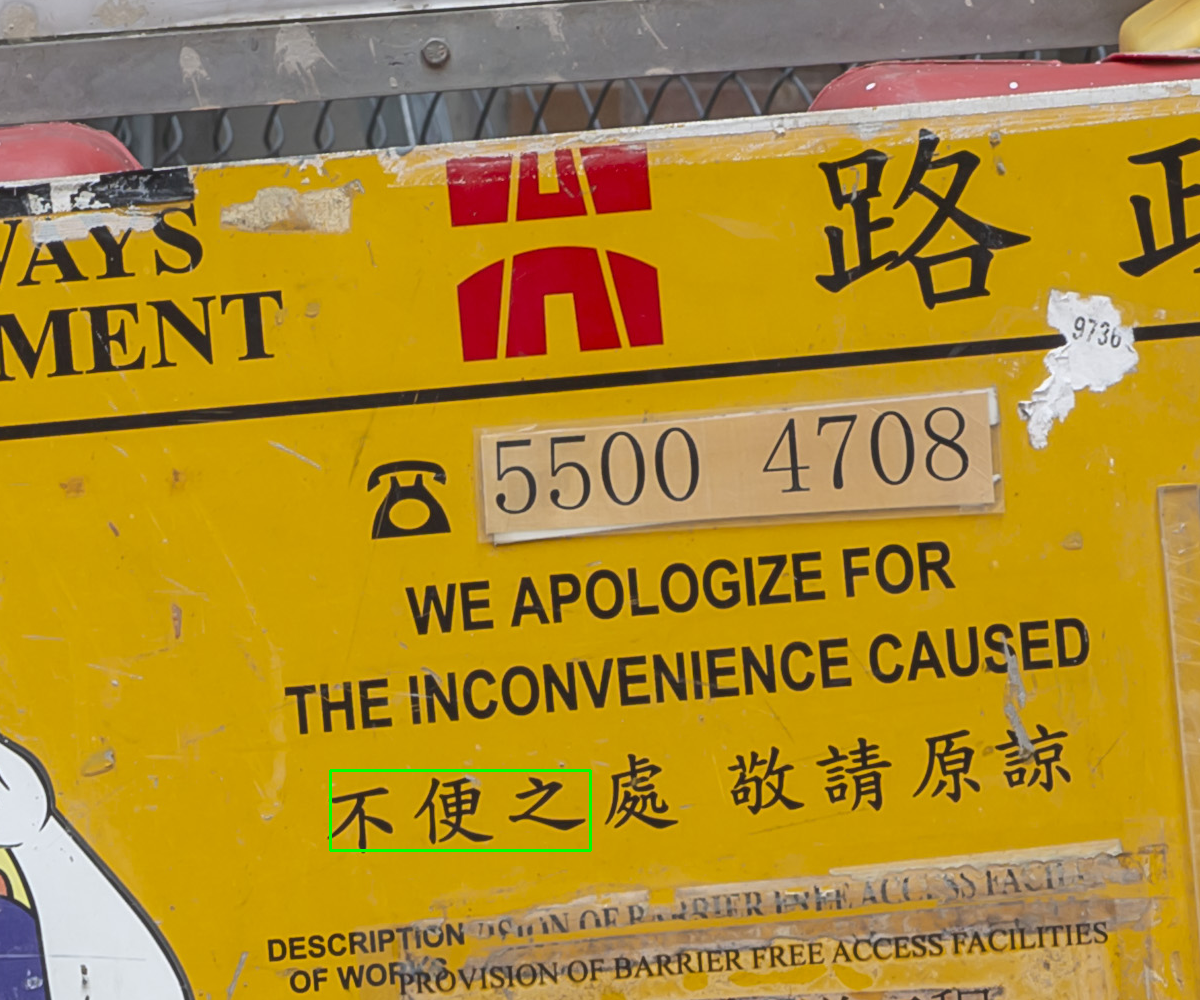} \\
					RealSRv3 ($\times 4$): \\
					Canon004 \\
				\end{tabular}
			\end{adjustbox}
			\hspace{-3mm}
			\begin{adjustbox}{valign=t}
				\begin{tabular}{ccc}
					\includegraphics[width=0.09\textwidth, height=0.05\textheight]{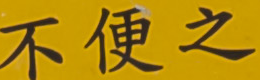} & 
					\hspace{-3mm}
                    \includegraphics[width=0.09\textwidth, height=0.05\textheight]{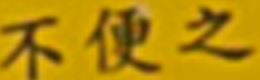} & 
					\hspace{-3mm}
					\includegraphics[width=0.09\textwidth, height=0.05\textheight]{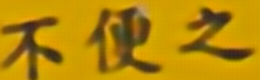} \\
					HR & \hspace{-3mm}
					Bicubic & \hspace{-3mm}
                    SRResNet~\cite{ledig2017photo}\\
					PSNR/SSIM & \hspace{-3mm}
                    26.06/0.7971 & \hspace{-3mm}
					26.50/0.8216 \\
					\includegraphics[width=0.09\textwidth, height=0.05\textheight]{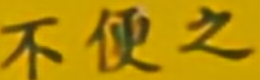} & 
					\hspace{-3mm}

                    \includegraphics[width=0.09\textwidth, height=0.05\textheight]{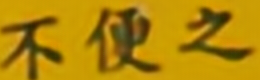} & 
					\hspace{-3mm}
					\includegraphics[width=0.09\textwidth, height=0.05\textheight]{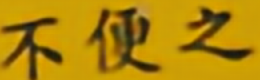} \\
					IMDN~\cite{hui2019lightweight} & \hspace{-3mm}
                    ESRT~\cite{lu2021efficient} & \hspace{-3mm}
					\textbf{Ours} \\
                    26.86/0.8388 & \hspace{-3mm}
					26.10/0.8166 & \hspace{-3mm}
					\textbf{27.31/0.8533} \\
				\end{tabular}
			\end{adjustbox}

            \begin{adjustbox}{valign=t}
				\begin{tabular}{c}
					\includegraphics[width=0.2\textwidth, height=0.125\textheight]{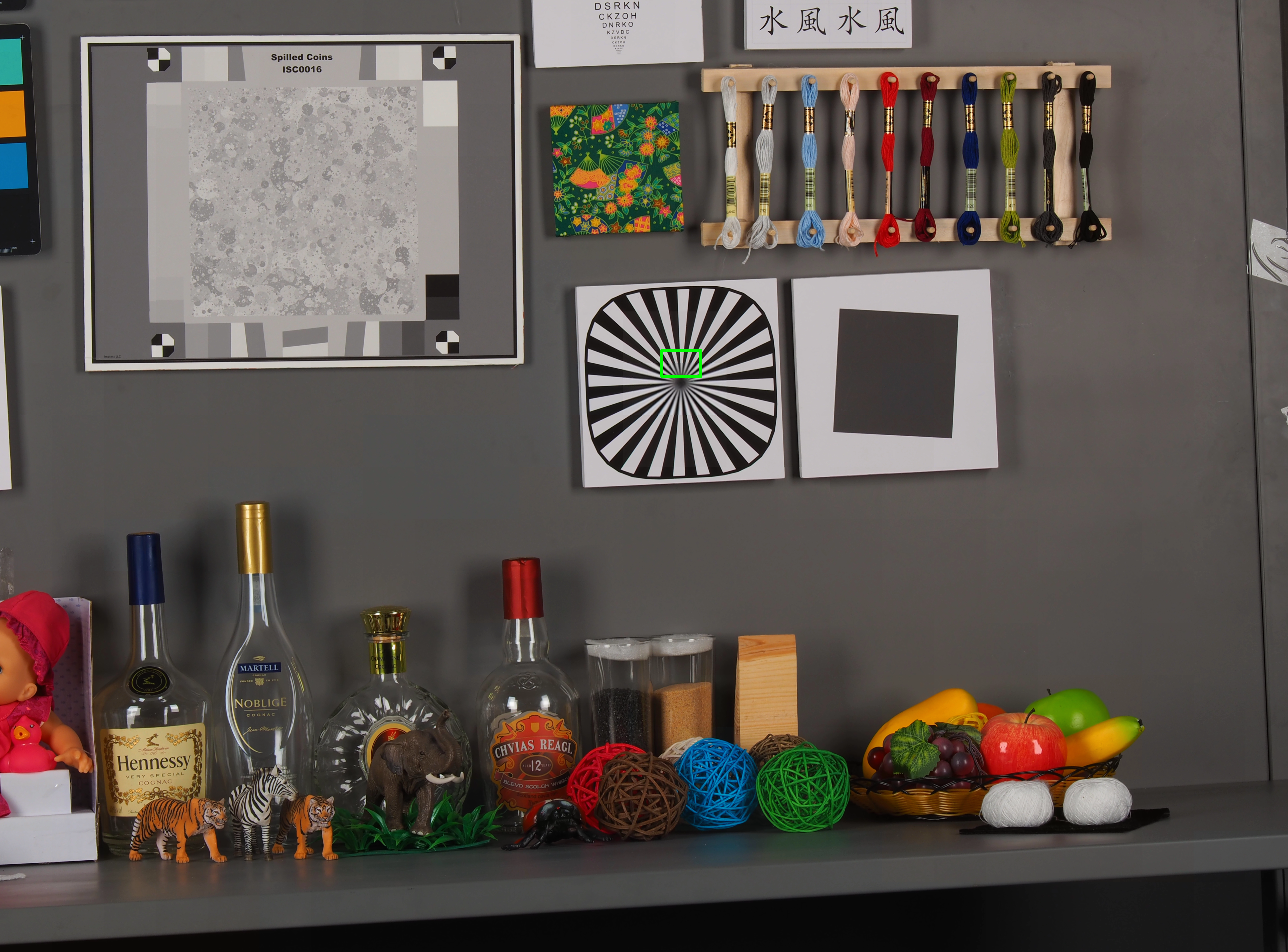} \\
					DRealSR ($\times 4$): \\
					P1160776 \\
				\end{tabular}
			\end{adjustbox}
			\hspace{-3mm}
			\begin{adjustbox}{valign=t}
				\begin{tabular}{ccc}
					\includegraphics[width=0.09\textwidth, height=0.05\textheight]{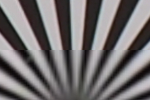} & 
					\hspace{-3mm}
                    \includegraphics[width=0.09\textwidth, height=0.05\textheight]{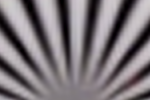} & 
					\hspace{-3mm}
					\includegraphics[width=0.09\textwidth, height=0.05\textheight]{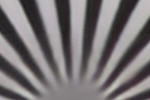} \\
					HR & \hspace{-3mm}
                    Bicubic & \hspace{-3mm}
					SRResNet~\cite{ledig2017photo} \\
					PSNR/SSIM & \hspace{-3mm}
                    30.26/0.9095 & \hspace{-3mm}
					31.86/0.9292 \\
					\includegraphics[width=0.09\textwidth, height=0.05\textheight]{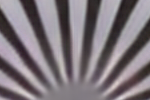} & 
					\hspace{-3mm}
                    \includegraphics[width=0.09\textwidth, height=0.05\textheight]{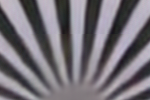} & 
					\hspace{-3mm}
					\includegraphics[width=0.09\textwidth, height=0.05\textheight]{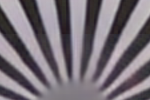} \\
                    IMDN~\cite{hui2019lightweight} & \hspace{-3mm}
					ESRT~\cite{lu2021efficient} & \hspace{-3mm}
					\textbf{Ours} \\
                    31.78/0.9299 & \hspace{-3mm}
					31.95/0.9312 & \hspace{-3mm}
					\textbf{32.22/0.9324} \\
				\end{tabular}
			\end{adjustbox}
			\\
			
	\end{tabular} }
	\caption{Visual comparisons on real-world datasets. (Including RealSRv3 and DRealSR).}
	\label{realsr-img}
\end{figure*}

\begin{table}[t]
	\centering
	\small
	\caption{Quantitative comparisons on real-world datasets.}
	\scalebox{0.83}{
	\begin{tabular}{|c|c|c|c|}
		\hline
		\multirow{2}{*}{Scale} & \multirow{2}{*}{Methods} & RealSRv3  & DRealSR\\
		\cline{3-4}
		& & PSNR / SSIM / LPIPS${\downarrow}$  & PSNR / SSIM / LPIPS${\downarrow}$  \\
		\hline
		\hline
		\multirow{4}{*}{$\times 4$} 
		& SRResNet~\cite{ledig2017photo}    & 27.70 / 0.7788 / 0.4010  & 31.18 / 0.8737 / 0.3618\\
    	& IMDN~\cite{hui2019lightweight}    & 27.76 / 0.7834 / 0.3766  & 31.30 / 0.8764 / 0.3411\\
        & ESRT~\cite{lu2021efficient}       & 27.61 / 0.7788 / 0.3895  & 31.17 / 0.8737 / 0.3556\\
        & CFIN    & \textbf{27.84} / \textbf{0.7872} / \textbf{0.3665}  & \textbf{31.40} / \textbf{0.8773} / \textbf{0.3383}\\
		\hline
	\end{tabular}
	}
	\label{tab:real-world}
	\vspace{-0.2cm}
\end{table}

\begin{table}[t]
	\centering
	\small
	\caption{Evaluate the effectiveness of masking mechanism.}
	\scalebox{1}{
	\begin{tabular}{|c|c|c|c|c|}
		\hline
		\multirow{2}{*}{Scale}  & \multirow{2}{*}{Mask}  & \multirow{2}{*}{Params} & \multirow{2}{*}{Multi-adds} & Set14  \\
		\cline{5-5}
		& & & & PSNR / SSIM   \\
		\hline
		\hline
		\multirow{2}{*}{$\times 4$} & \XSolid & 186.8K   &6.18220G & 28.22 / 07723\\
		
		   & \Checkmark         & 187.9K   &6.21538G     & \textbf{28.30} / \textbf{0.7738} \\
		\hline
	\end{tabular}
	}
	\label{tab:RIFU}
\end{table}

\begin{table}[t]
	\centering
	\small
	\caption{Evaluate the effectiveness of Gumbel-Softmax.}
	\scalebox{1}{
	\begin{tabular}{|c|c|c|c|c|}
		\hline
		\multirow{2}{*}{Scale} & \multirow{2}{*}{Modules}  & \multirow{2}{*}{Params} & \multirow{2}{*}{Multi-adds} & Set5  \\
		\cline{5-5}
		& & & & PSNR / SSIM   \\
		\hline
		\hline
		\multirow{3}{*}{$\times 4$} 
		& RIFU-Maxpool  & 165K   & 4.89G & 31.70 / 0.8876\\
		& RIFU-Softmax                & 165K   & 4.89G   & 31.71 / 0.8877 \\
		& RIFU         & 165K   & 4.89G     & \textbf{31.75} / \textbf{0.8886} \\
		\hline
	\end{tabular}
	}
	\label{tab:Gumbel-Softmax}
\end{table}

\subsection{Comparison with Advanced Lightweight SISR Models}
In this section, we compare our proposed CFIN with other advanced lightweight SISR models to verify the effectiveness of the proposed model. In addition, we provide a version with the self-integrating strategy~\cite{timofte2016seven} and denote it as CFIN+. In TABLE~\ref{tab:sota}, we compare CFIN with CNN-based models. From the table, we can clearly observe that our CFIN+ and CFIN stand out from these methods and achieve the best and the second-best results on almost all datasets. It is worth mentioning that our CFIN consumes fewer parameters and computations than most methods. This benefits from the well-designed CNN and Transformer in CFIN, which can efficiently extract the local features and integrate the global information of the image. To further demonstrate the superiority of CFIN, in TABLE~\ref{tab:swinir}, we also provide a comprehensive comparison with some advanced Transformer-based models, including the lightweight version of SwinIR*~\cite{liang2021swinir}, ESRT~\cite{lu2021efficient}, and LBNet~\cite{gao2022lightweight}. All these models are the most advanced lightweight SISR models. From TABLE~\ref{tab:swinir}, we can see that our CFIN achieves better results than ESRT and LBNet with fewer parameters and computation. Compared with SwinIR*, our CFIN can still achieve close results than it with fewer parameters, Multi-adds, and execution time. It is worth mentioning that SwinIR* uses a pre-trained model for initialization, and sets the patch size as $64 \times 64$ during training. Extensive experiments have shown that the larger the patch size, the better the results. Meanwhile, some previous works~\cite{lim2017enhanced, li2020mdcn} have pointed out that the performance of models trained with multiple upsampling factors shows better results than the single one since using the inter-scale correlation between different upsampling factors can improve the model performance. Therefore, these methods can further improve the model's performance. Moreover, we provide the results of CFIN-L, the larger version of CFIN. It can be seen that the results of CFIN-L even surpass SwinIR* on some datasets and still keep fewer parameters and less time. This is due to the rational structural design and strong modeling capabilities of CFIN. All these results fully illustrate the strong competitiveness of CFIN in balancing the model size and performance. In addition, we also provide a visual comparison of CFIN with other SISR methods in Fig.~\ref{visual}. Our CFIN can reconstruct high-quality images with more accurate textures details and edges. This further demonstrates the effectiveness of the proposed CFIN.

\begin{table}[t]
	\small
	\centering
	\caption{Evaluate the effectiveness of CIAM.}
	\scalebox{0.9}{
	\begin{tabular}{|l|c|c|c|c|c|}
		\hline
		Modules & Params & Multi-adds & Set5 & Set14  & U100\\
		\hline
		\hline
		CFIN+RCAB~\cite{zhang2018image} & 192K &17.35G & 31.63 & 28.19  & 25.29 \\
		CFIN+IMDB~\cite{hui2019lightweight} & 148K &8.47G & 31.54 & 28.19  & 25.23 \\
		CFIN+RFDB~\cite{liu2020residual} & 140K &7.77G & 31.68 & 28.26  & 25.34 \\
		CFIN+LB~\cite{luo2020latticenet}      & 145K &8.16G & 31.66 & 28.22  & 25.33 \\
		CFIN+HPB~\cite{lu2021efficient}   & 150K &8.91G & 31.59 & 28.21  & 25.32 \\
		CFIN+WDIB~\cite{gao2022feature} & 147K &5.26G & 31.68 & 28.26  & 25.37 \\
		CFIN+CIAM (Ours) & 188K & 6.22G & \textbf{31.87} & \textbf{28.30}  & \textbf{25.44} \\
		\hline
	\end{tabular}}
	\label{tab:CIAM}
\end{table}

\begin{figure}[t]
\centering
\includegraphics[width=0.48\textwidth]{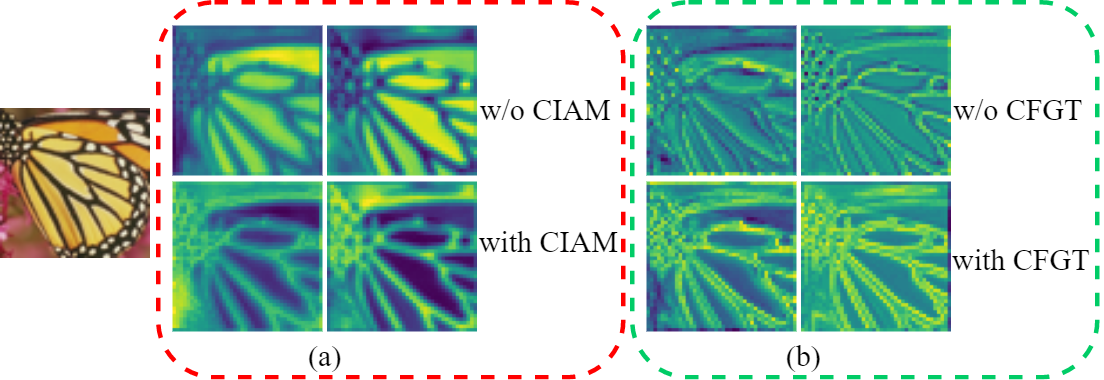}
\caption{Visual comparison of features with or without CIAM and CFGT. CIAM is mainly composed of three RIFUs, and it serves to remove redundant features from the image.}
\label{feature-map}
\end{figure}

\subsection{Real-World Image Super-Resolution}
To validate the performance of our proposed method in real-world scenarios, we compare our CFIN with several classical models on RealSRv3 dataset~\cite{cai2019toward} and DRealSR dataset~\cite{wei2020component}. These methods include SRResNet~\cite{ledig2017photo}, IMDN~\cite{hui2019lightweight}, and ESRT~\cite{lu2021efficient}. To speed up the training, we uniformly train all methods using a patch size of 24$ \times $24. It is worth noting that except for the conventional PSNR/SSIM index, we also provide the visual sensory index LPIPS for better comparisons. As can be seen from TABLE~\ref{tab:real-world}, our method achieves the best performance for all metrics on both datasets. In addition, we give visual comparisons in Fig.~\ref{realsr-img}. Both results on the DRealSR dataset and the RealSRv3 dataset have demonstrated that our method has a good recovery effect on textural features such as text and line segments. This further demonstrates that our proposed CFIN is also effective on real datasets.

\section{Ablation Studies}~\label{AS}

\subsection{Network Investigations}
\subsubsection{The effectiveness of RIFU}

In RIFU, we use feature masks to remove redundant features. To verify the effectiveness of this mechanism, we remove the mask and provide the results in TABLE~\ref{tab:RIFU}. According to the table, we can see that the PSNR value increases by 0.08dB after using the masking mechanism under the slight increase in the number of parameters. This effectively illustrates the effectiveness of the mechanism. As we know, the max-pooling operation can filter redundant features, and softmax can predict the probability distribution of different features. These functions also can make the model focus on the main features. To prove the effectiveness of the Gumbel-Softmax (GS) in RIFU, we replace the Gumbel-Softmax function with the max-pooling operation and softmax function, respectively. According to TABLE~\ref{tab:Gumbel-Softmax}, we can see that the performance of the models will drop from 31.75dB to 31.71dB when only using the softmax function, and the performance will drop to 31.70dB when using the max-pooling operation. Therefore, we choose the Gumbel-Softmax operation in the final model. The above experiments verify the effectiveness and necessity of each mechanism in RIFU.

Furthermore, to understand what part of the image our proposed RIFU is focusing on, we visualize the feature maps for the different layers of the model. Fig.~\ref{feature-map} (a) shows the feature map comparison of the presence or absence of CIAM in the model. It is worth noting that CIAM is composed of three RIFUs. As can be seen from the figure, when CIAM is not present in the model, the model focuses more on the flat areas between the butterfly textures. However, the areas that determine the visual quality of the image are the critical texture features, and such a focus is not good enough to improve the quality of reconstructed images. On the contrary, when the CIAM is introduced in the model, the attention of the model shifts to complex texture features. This proves that our proposed RIFU enables the model to focus more on potentially useful information.

\subsubsection{The effectiveness of CIAM}
To verify the effectiveness of CIAM, we replace CIAM with some commonly used feature extraction blocks in lightweight SISR models. It is worth noting that we remove the Transformer stage in each CT block to speed up the training process. According to TABLE~\ref{tab:CIAM}, we can find that when the model uses CIAM for feature extraction, the model achieves the best results at the expense of a small increase in the number of parameters. This fully demonstrates the effectiveness of the proposed CIAM.

\subsubsection{The effectiveness of CGM} 
Context Guided MaxConv is the basic unit in our Transformer stage, which is responsible for reasoning and guiding the entire network. Compared with the two basic units of convolutional and linear layers, our CGM can adjust the weights adaptively and select reasonable contextual information for reconstruction, which is not available in convolutional or linear units. To verify its importance, we replace CGM with the linear layer and group convolution, respectively. Meanwhile, we set their parameters as close as possible. According to TABLE~\ref{tab:CGM}, we can see that our proposed CGM achieves better performance with fewer parameters and Multi-adds. This fully validates the excellence and effectiveness of the proposed CGM.


\subsubsection{The effectiveness of CFGT}
\begin{figure}[t]
\centering
\includegraphics[width=0.45\textwidth]{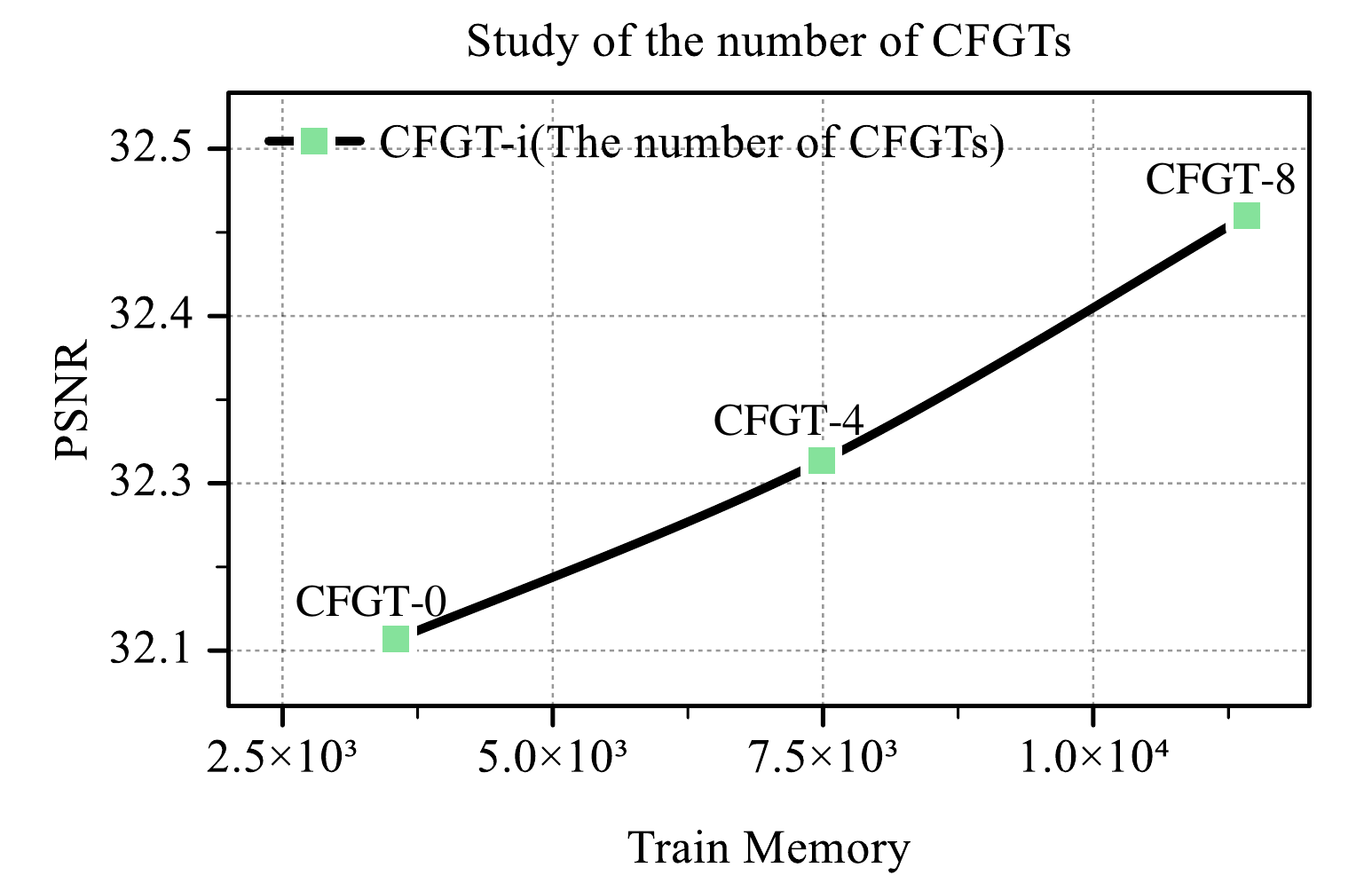}
\caption{Efficiency trade-off of CFGT on Set5($ \times 4$).}
\label{Efficiency}
\end{figure}

\begin{figure}[t]
\centering
\includegraphics[width=0.48\textwidth]{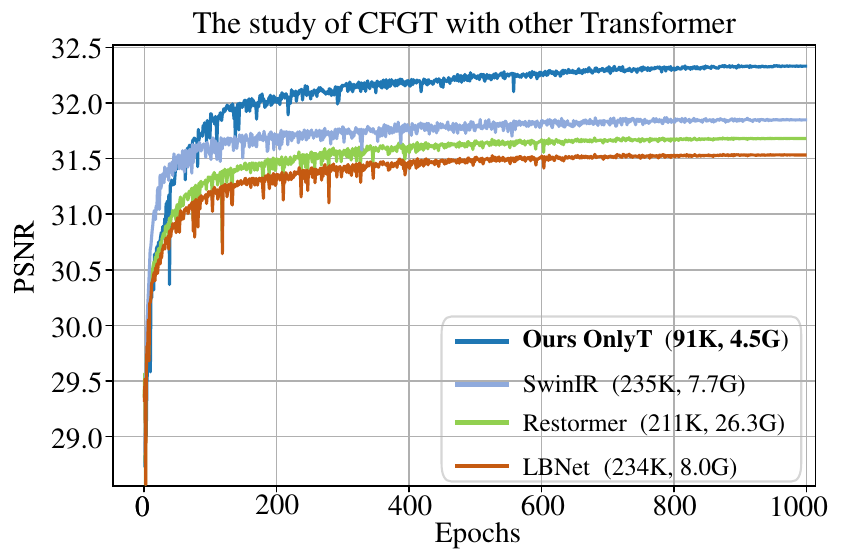}
\caption{Performance comparison of CFGT with other Transformer modules on Set5($ \times 4$).}
\label{compare_CFGT}
\end{figure}

Compared with traditional Transformers, CFGT has two main differences in structure: the first is K, V for contextual communication; the second is cross-scale information exchange brought by CGM with different receptive fields. To assess the effectiveness of such a design, we provide a set of ablation experiments in TABLE~\ref{tab:CFGT}, where \textbf{KV} represents the interaction of vectors $K$ and $V$ between the upper and lower CGA, and \textbf{Cross} represents the cross-receptive fields mechanism brought about by setting $K$ of different sizes in CGA. It can be seen that our architectural design solution can significantly improve the performance of the model with almost no additional computational cost. Meanwhile, we make an efficiency trade-off study on the number of CFGTs. In Fig.~\ref{Efficiency}, CFGT-N indicates that the model contains $N$ CFGTs while keeping the CNN part unchanged. It can be seen that the model with 8 CFGTs has the best performance, so we select 8 CFGTs in CFIN. In TABLE~\ref{tab:8CFGTs}, OnlyT indicates that only the Transformer part of the CFIN is retained. Compared with other advanced lightweight SISR models, it can be seen that our CFGT achieves promising results with only 91K parameters. This verifies the effectiveness and advancement of CFIN. In Fig.~\ref{compare_CFGT}, we also compare OnlyT with several Transformer-based SISR methods. It is worth noting that we only choose the Transformer part of these methods and use a small model for ablation studies. As can be seen from Fig.~\ref{compare_CFGT}, our approach has better performance with less computational consumption.

\begin{table}[t]
	\small
	\centering
	\caption{Evaluate the effectiveness of CGM.}
	\scalebox{0.9}{
	\begin{tabular}{|l|c|c|c|c|c|}
		\hline
		Modules & Params & Multi-adds & Set5 & B100 & U100\\
		\hline
		\hline
		CFIN+GConv & 197K &5.7981G & 31.95   & 27.41 & 25.68 \\
		CFIN+Linear  & 198K &4.1104G & 29.19   & 26.37 & 23.62 \\
		CFIN+CGM (Ours) & 178K & 4.1105G & \textbf{32.14}  & \textbf{27.43} & \textbf{25.71} \\
		\hline
	\end{tabular}
	}
	\label{tab:CGM}
\end{table}

\begin{table}[t]
	\centering
	\small
	\caption{Evaluate the effectiveness of the contextual interaction mechanism within CFGT. }
	\scalebox{0.98}{
	\begin{tabular}{|c|c|c|c|c|c|}
		\hline
		\multirow{2}{*}{Scale} & \multirow{2}{*}{KV} & \multirow{2}{*}{Cross}  & \multirow{2}{*}{Params} & \multirow{2}{*}{Multi-adds} & Set5  \\
		\cline{6-6}
		& & & & & PSNR / SSIM   \\
		\hline
		\hline
		\multirow{4}{*}{$\times 4$} & \XSolid & \XSolid & 177.5K   &4.1104G & 31.85 / 0.8900\\
		& \XSolid & \Checkmark & 177.8K   &4.1105G & 32.05 / 0.8928\\
		& \Checkmark   & \XSolid                & 177.5K   &4.1104G   & 31.93 / 0.8904 \\
		& \Checkmark   & \Checkmark         & 177.8K   &4.1105G     & \textbf{32.14} / \textbf{0.8940} \\
		\hline
	\end{tabular}
	}
	\label{tab:CFGT}
\end{table}

\begin{table}[t]
	\small
	\centering
	\caption{Comparison of CFGT with other methods.}
	\scalebox{0.85}{
	\begin{tabular}{|l|c|c|c|c|c|c|c|}
		\hline
		Modules & Params & Multi-adds & Set5 & Set14 & B100 & U100\\
		\hline
		\hline
		PAN~\cite{zhao2020efficient}      & 272K &28.2G & 32.13 & 28.61 & \textbf{27.59} & \textbf{26.11} \\
		MAFFSRN~\cite{muqeet2020multi}    & 441K &19.3G & 32.18 & 28.58 & 27.57 & 26.04 \\
		RFDN~\cite{liu2020residual}       & 550K &31.6G & 32.24 & 28.61 & 27.57 & 26.11 \\
		FDIWN-M~\cite{gao2022feature}     & 454K &19.6G & 32.17 & 28.55  & 27.58  & 26.02\\
		OnlyT (Ours)                     & 91K  & 4.5G & \textbf{32.33} & \textbf{28.61} & 27.58  & 26.08 \\
		\hline
	\end{tabular}}
	\label{tab:8CFGTs}
\end{table}

Furthermore, Fig.~\ref{feature-map} (b) illustrates the effect of the presence or absence of CFGT in the model on the area of concern of the model. It can be seen from the figure that the model only focuses on the part of the contour texture when there is no CFGT, while the model deepens its focus on the contour texture when there is CFGT, and also focuses on some of the pixel points around the contour. This also demonstrates that our proposed CFGT can be combined with contextual information on important image regions for image reconstruction. These experiments all validate the effectiveness of the proposed CFGT which can better convey contextual information.



\subsubsection{Visual analysis of CT Block}
\begin{figure}[t]
\centering
\includegraphics[width=0.48\textwidth]{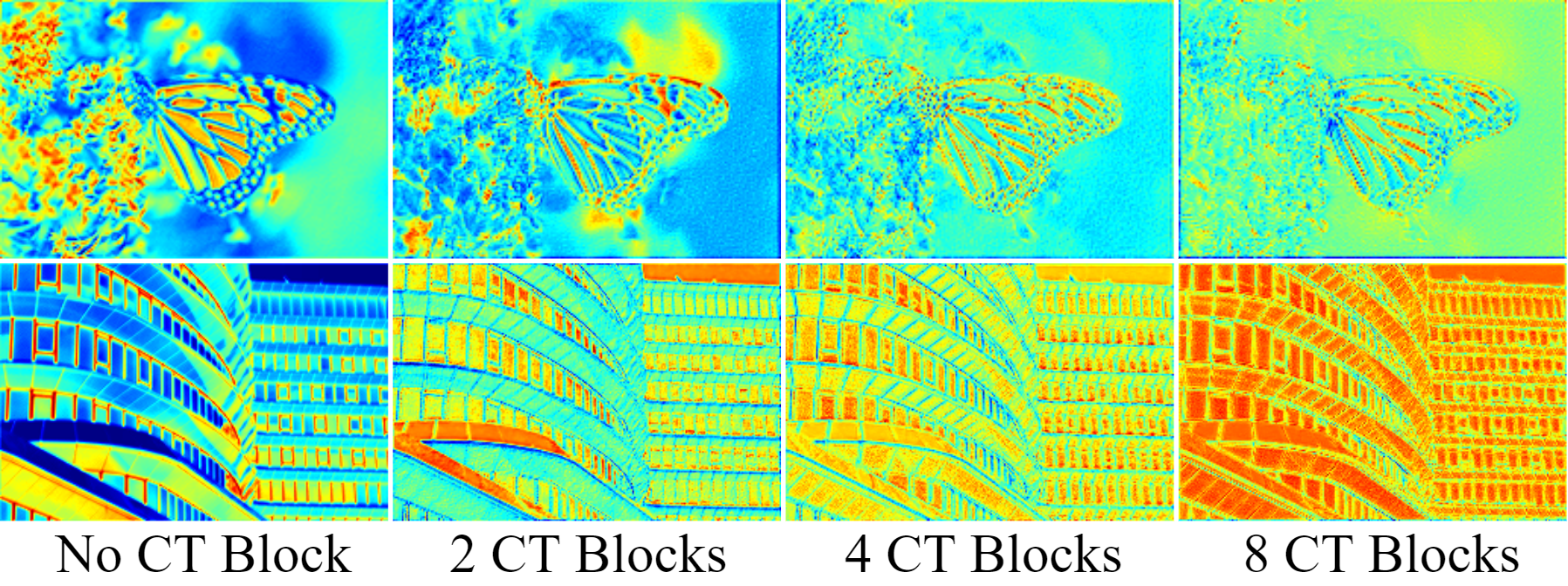}
\caption{The visualization of different numbers of CT blocks.}
\label{visual CFGT}
\end{figure}

To explore which regions of the image the CT Block that CNN combines with Transformer will focus on, we present the visual heatmaps of different numbers of CT Blocks in Fig.~\ref{visual CFGT}. Among them, the color close to red in the image represents the part where the attention is focused. So we can see that in the absence of a CT Block, the attention is only focused on a small number of texture features, and as the number of CT Blocks increases, more fine-grained features in images are paid attention to. It also indicates that CT Blocks allow the model to progressively focus on the surrounding fine-grained pixels that the model needs to reconstruct the texture when recovering important texture features. This means that more CT Blocks make the model tends to recover more accurate detailed features, which is conducive to image restoration.

\begin{table*}[t]
	\centering
	\small
	\caption{Evaluate the feasibility of combining CNN with Transformer.}
	\label{tab:combining}
	\scalebox{0.9}{
	\begin{tabular}{|c|l|c|c|c|c|c|c|c|c|c|}
		\hline
		\multirow{2}{*}{Scale} & \multirow{2}{*}{Methods} & \multirow{2}{*}{Params} & \multirow{2}{*}{GPU Memory}  & \multirow{2}{*}{Time}  & Set5 & Set14 & BSD100 & Urban100 & Manga109   \\
		\cline{6-10}
		& & & & &PSNR / SSIM & PSNR / SSIM & PSNR / SSIM & PSNR / SSIM & PSNR / SSIM  \\
		\hline
		\hline
		\multirow{3}{*}{$\times 4$} &Pure-CNN & 720K &3845M & 0.016s   & 32.12 / 0.8939 & 28.52 / 0.7796  &  27.52 / 0.7343 & 25.96 / 0.7810 & 30.29 / 0.9057 \\
		&Pure-Transformer & 94K &11549M & 0.038s   & 32.48 / 0.8980 & 28.69 / 0.7835 &  27.64 / 0.7385 & 26.24 / 0.7896 & 30.62 / 0.9102\\
		&CFIN (Ours) & 699K &11453M & 0.035s   & \textbf{32.49} / \textbf{0.8985} & \textbf{28.74} / \textbf{0.7849} &  \textbf{27.68} / \textbf{0.7396} & \textbf{26.39} / \textbf{0.7946} & \textbf{30.73} / \textbf{0.9124} \\
		\hline
	\end{tabular}
	}
\end{table*}

\begin{figure}[t]
 \centerline{\includegraphics[width=8cm,trim=0 0 30 40]{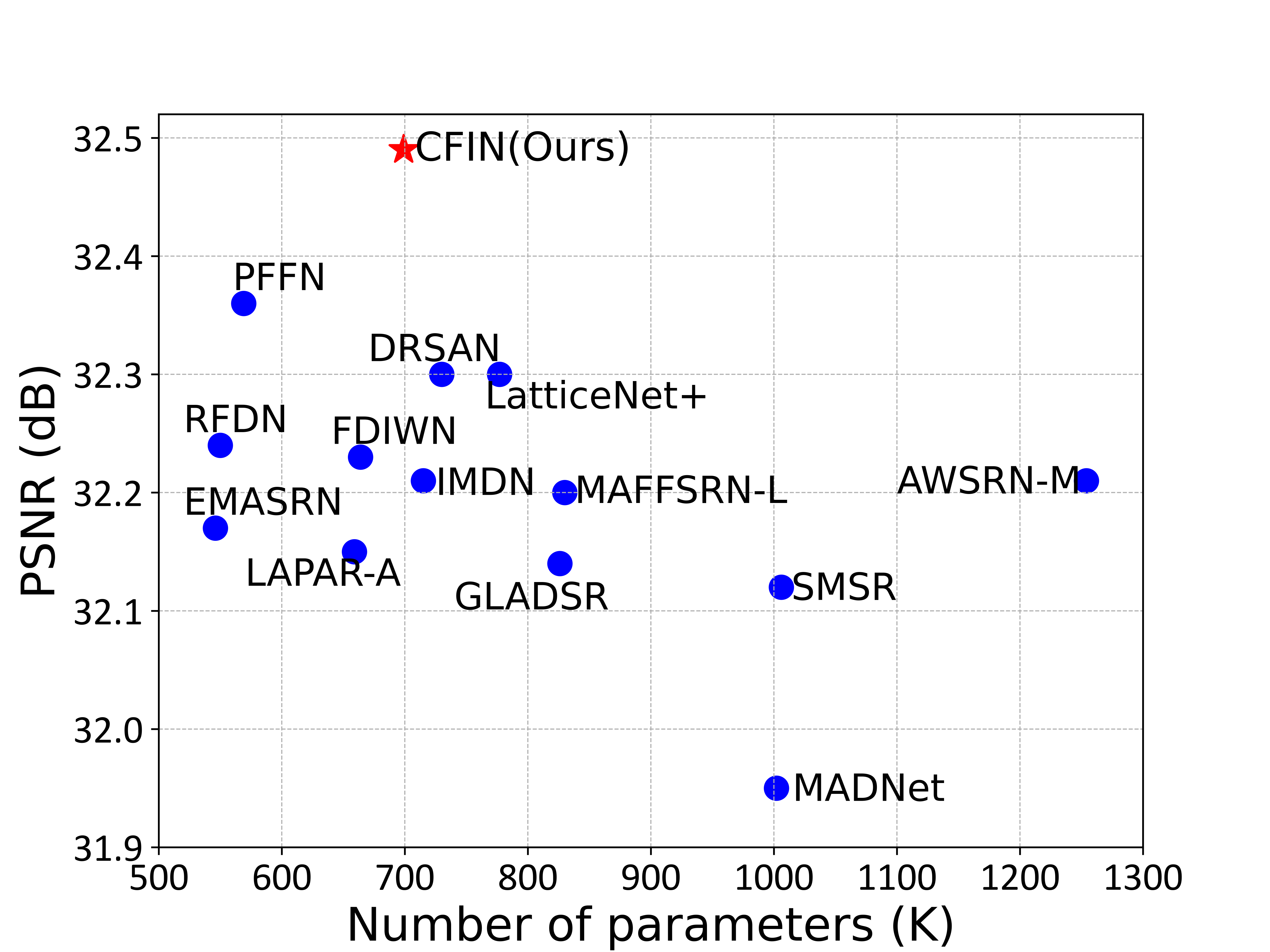}}
 \caption{Model performance and size comparison on Set5 ($\times 4$). Obviously, our CFIN achieves the best balance between model performance and size.}
 \label{CFIN_Tradeoff_Params}
\end{figure}

\begin{figure}[!t]
 \centerline{\includegraphics[width=8cm,trim=0 0 30 40]{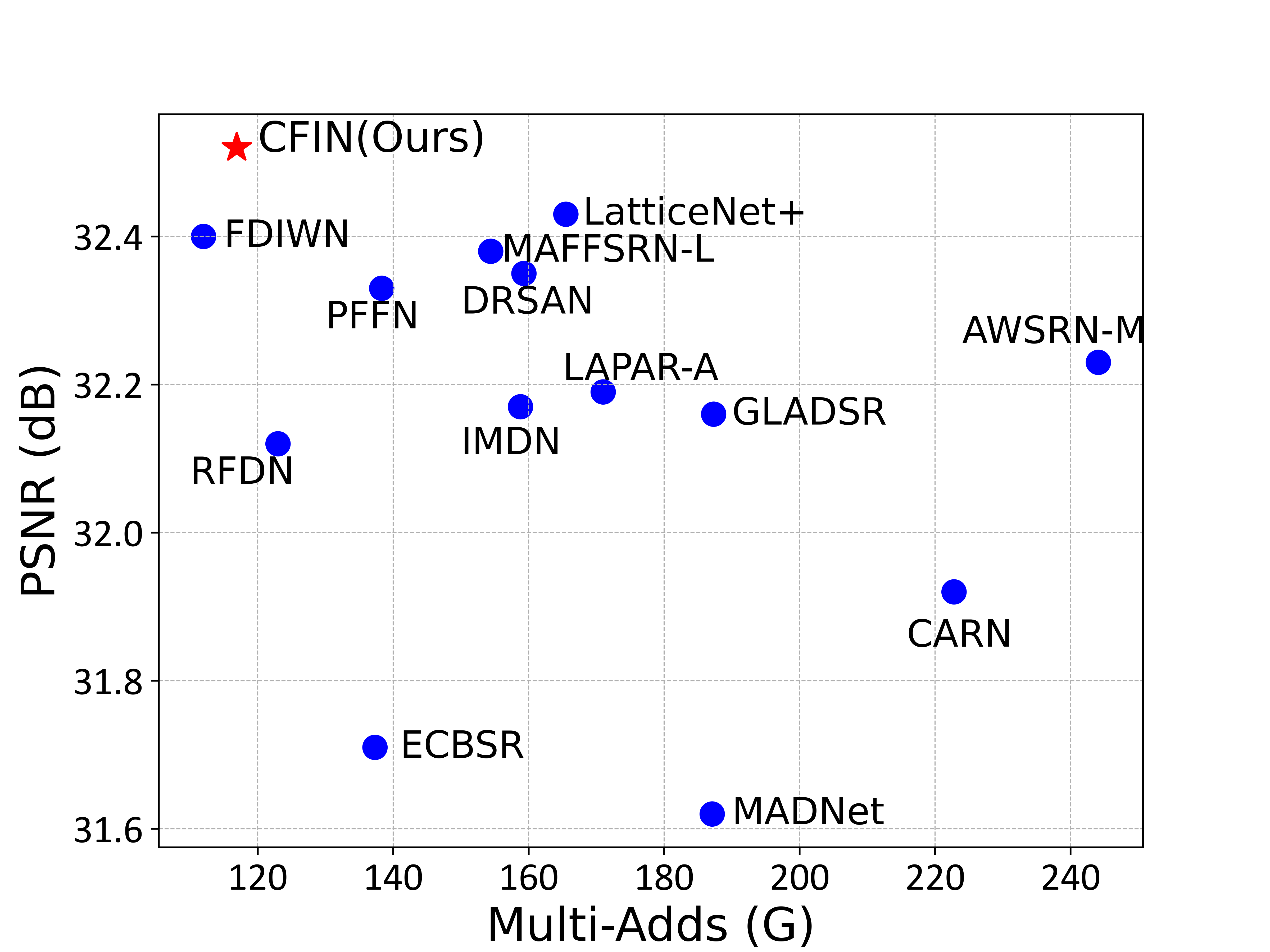}}
 \caption{Model performance and Multi-Adds comparison on Urban100 ($\times 2$). Obviously, our CFIN achieves the best balance between model performance and Multi-Adds.}
 \label{CFIN_Tradeoff_MultAdds}
\end{figure}

\subsection{Complementarity of CNNs and Transformers}
CNN can be used to extract local features, and Transformer has powerful global modeling capabilities, both of which are crucial for high-quality image restoration. In TABLE~\ref{tab:combining}, we provide a comparison of CFIN with Pure-CNN and Pure-Transformer versions. Among them, Pure-CNN and Pure-Transformer represent variant models with only the CNN part or the Transformer part, respectively. For a fair comparison, the number of parameters of Pure-CNN is set as close to CFIN as possible, and the memory consumption of Pure-Transformer is set as close to CFIN as possible. It can be seen from the table that neither Pure-CNN nor Pure-Transformer cannot achieve the performance of the original CFIN. When using CNN alone, it can reduce the consumption of GPU memory, but it is difficult to improve the performance of the model even with more parameters. When using the Transformer alone, the parameters of the model can be greatly reduced, but the consumption of GPU memory will rise rapidly, and its performance still hardly exceeds the original CFIN, which is not conducive to the practical application of the model. Therefore, we chose the hybrid model of CNN and Transformer, which can achieve a good balance between the size, GPU memory consumption, and performance of the model. Therefore, we can draw a conclusion that CNN and Transformer are complementary, and the combination of these two parts is feasible.

\subsection{Model Complexity Analysis}
We present the trade-off between our CFIN and other advanced SISR models in terms of PSNR, parameter amount, and inference time in Fig.~\ref{Time}. Obviously, CFIN attains the best performance among models with similar execution times and achieves the best balance in model complexity, inference time, and performance. 

In Fig.~\ref{CFIN_Tradeoff_Params} and Fig.~\ref{CFIN_Tradeoff_MultAdds}, we also provide the parameter, Multi-Adds, and performance comparisons of CFIN with other advanced SISR models. It can be seen that our CFIN also achieved the best PSNR results under the premise of comparable calculations. Therefore, we can draw a conclusion that our CFIN is a lightweight and efficient model, which achieves the best balance between the model size and performance.

\section{Discussion}~\label{DS}
In the proposed model, we use some matrix multiplication in our model, so the training memory is slightly larger. However, this does not mean the proposed method is meaningless. Because our method, like most other methods, can be trained and tested on a single NVIDIA RTX 2080Ti GPU and has fewer computational costs and faster inference during testing, it is friendly to the deployment of the model on the mobile device. We also note that some researchers~\cite{lu2021efficient} claimed that their methods can greatly reduce the memory required for Transformer training. In addition, due to the fact that the original image details are severely corrupted under a large sampling factor, our method still can not recover the fine edge texture details well. Fortunately, we also note that partly pre-training-based methods~\cite{zhang2018residual,li2020mdcn} are able to use prior knowledge to mitigate the above problem. In our future works, We will further explore the effectiveness of these strategies and introduce them into CFIN to further improve the model.

\section{Conclusions}~\label{CL}
In this paper, we proposed a lightweight and efficient Cross-receptive Focused Inference Network (CFIN) for SISR. The network consists of sequentially cascaded CT Blocks, each composed of a Cross-scale Information Aggregation Module (CIAM) and a Cross-receptive Field Guide Transformer (CFGT). By removing redundant features and combining contextual information to dynamically perform image restoration, our method can effectively fuse the advantages of CNN and Transformer. Extensive experiments have shown that our CFIN can effectively combine contextual information for fine-grained learning, which strike a good balance between the performance and complexity of the model and outperform existing state-of-the-art methods.

\bibliographystyle{IEEEtran}
\bibliography{sample-base}

\end{document}